\documentclass{article}

\usepackage{microtype}
\usepackage{graphicx}
\usepackage{booktabs} 
\usepackage{float}
\usepackage{hyperref}
\usepackage{diagbox}
\usepackage{stfloats}

\usepackage[accepted]{icml2023}

\usepackage{amsmath}
\usepackage{amssymb}
\usepackage{mathtools}
\usepackage{amsthm}
\usepackage[export]{adjustbox}
\usepackage{xcolor}
\usepackage[capitalize,noabbrev]{cleveref}

\theoremstyle{plain}

\theoremstyle{definition}

\theoremstyle{remark}

\usepackage[textsize=tiny]{todonotes}

\usepackage{subcaption}
\usepackage{tabulary}

\newcommand{\mat}[1]{\mathbf{#1}}
\newcommand{\grad}[1]{\nabla_{#1}}

\icmltitlerunning{Universal Guidance for Diffusion Models}

\begin{document}
\icmltitlerunning{Universal Guidance for Diffusion Models\hfill\thepage}
\twocolumn[
\icmltitle{Universal Guidance for Diffusion Models}

\icmlsetsymbol{equal}{*}

\begin{icmlauthorlist}
\icmlauthor{Arpit Bansal}{equal,umd}
\icmlauthor{Hong-Min Chu}{equal,umd}
\icmlauthor{Avi Schwarzschild}{umd}
\icmlauthor{Soumyadip Sengupta}{unc}
\icmlauthor{Micah Goldblum}{nyu}
\icmlauthor{Jonas Geiping}{umd}
\icmlauthor{Tom Goldstein}{umd}
\end{icmlauthorlist}

\icmlaffiliation{umd}{Department of Computer Science, University of Maryland, College Park}
\icmlaffiliation{nyu}{Department of Computer Science, New York University}
\icmlaffiliation{unc}{Department of Computer Science, University of North Carolina at Chapel Hill}

\icmlcorrespondingauthor{Arpit Bansal}{bansal01@umd.edu}
\icmlcorrespondingauthor{Hong-Min Chu}{hmchu@umd.edu}

\icmlkeywords{Machine Learning, ICML}

\vskip 0.3in

]



\printAffiliationsAndNotice{\icmlEqualContribution} 



\begin{abstract}
Typical diffusion models are trained to accept a particular form of conditioning, most commonly text, and cannot be conditioned on other modalities without retraining. 
%
In this work, we propose a universal guidance algorithm that enables diffusion models to be controlled by arbitrary guidance modalities without the need to retrain any use-specific components.
We show that our algorithm successfully generates quality images with guidance functions including segmentation, face recognition, object detection, and classifier signals. Code is available at \href{https://github.com/arpitbansal297/Universal-Guided-Diffusion}{github.com/arpitbansal297/Universal-Guided-Diffusion}.


\end{abstract}



\section{Introduction}

\begin{figure}[h!]



%
    \includegraphics[width=0.46\textwidth] {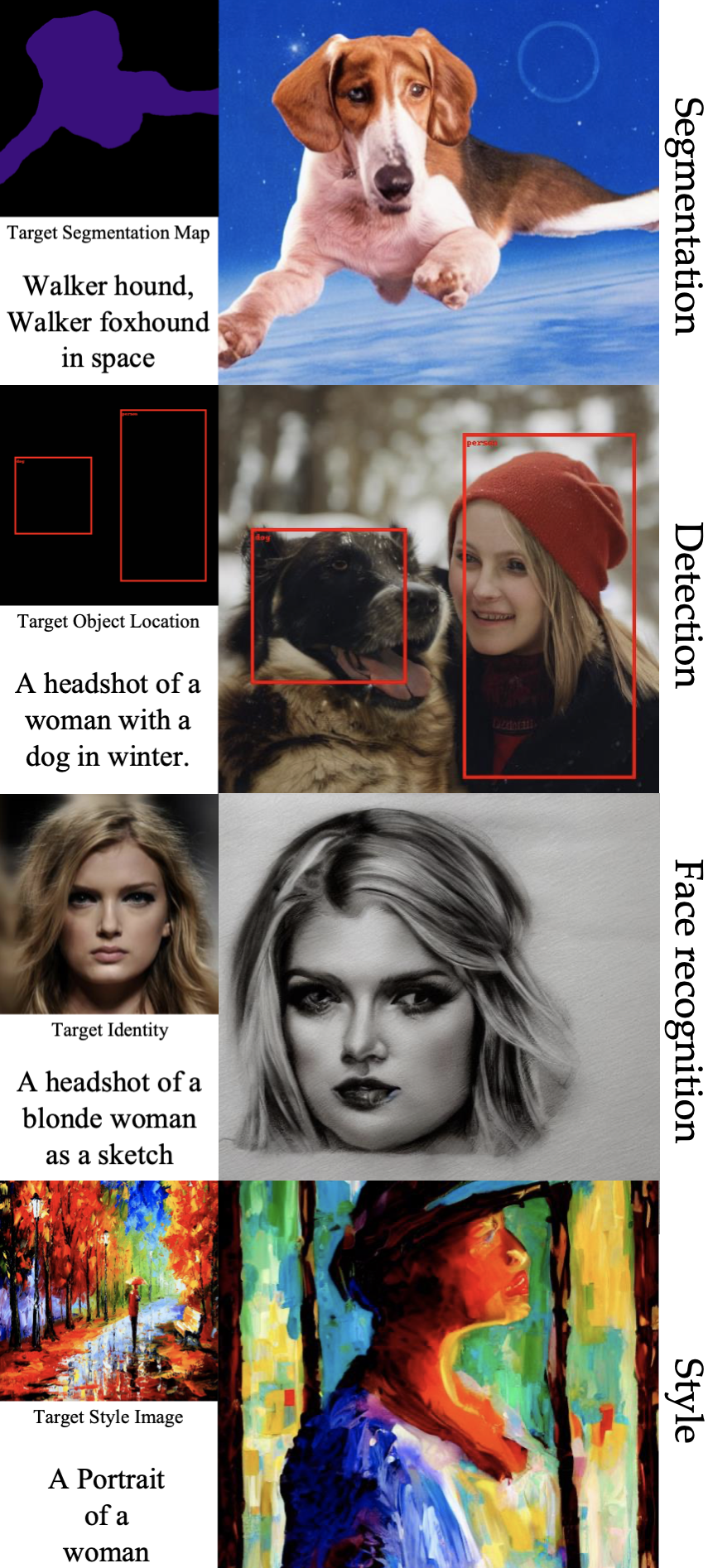}
    \label{fig:cover}
    \caption{\looseness -1 Diffusion guided by off-the-shelf networks.}
    \vspace{-.3cm}
\end{figure}

Diffusion models are powerful tools for creating digital art and graphics.  Much of their success stems from our ability to carefully control their outputs, customizing results for each user's individual needs.  Most models today are controlled through {\em conditioning}. With conditioning, the diffusion model is built from the ground up to accept a particular modality of input from the user, be it descriptive text, segmentation maps, class labels, etc.  
While conditioning is a powerful tool, it results in models that are handcuffed to a single conditioning modality.  If another modality is required, a new model needs to be trained, often from scratch. Unfortunately, the high cost of training makes this prohibitive for most users. 

A more flexible approach to controlling model outputs is to use {\em guidance}.  In this approach, the diffusion model acts as a generic image generator, and is not required to understand a user's instructions. The user pairs this model with a guidance function that measures whether some criterion has been met.  For example, one could guide the model to minimize the CLIP score between the generated image and a text description of the user's choice.
During each iteration of image creation, the iterates are nudged down the gradient of the guidance function, causing the final generated image to satisfy the user's criterion. 

In this paper, we study guidance methods that enable any off-the-shelf model or loss function to be used as guidance for diffusion. Because guidance functions can be used without re-training or modification, this form of guidance is {\em universal} in that it enables a diffusion model to be adapted for nearly any purpose.  

From a user perspective, guidance is superior to conditioning, as a {\em single} diffusion network is treated like a foundational model that provides universal coverage across many use cases, both commonplace and bespoke. Unfortunately, it is widely believed that this approach is infeasible.  While early diffusion models relied on classifier guidance~\cite{dhariwal21diffusion_beats_gan}, the community quickly turned to classifier-free schemes~\cite{ho2022classifierfree} that require a model to be trained from scratch on class labels with a particular frozen ontology that cannot be changed~\cite{nichol2021glide,rombach2022stablediff,bansal2022colddiff}.

The difficulty of using guidance stems from the domain shift between the noisy images used by the diffusion sampling process and the clean images on which the guidance models are trained. When this gap is closed, guidance can be performed successfully.  For example, \citet{nichol2021glide} successfully use a CLIP model as guidance, but only after re-training CLIP from scratch using noisy inputs.  Noisy retraining closes the domain gap, but at a very high financial and engineering cost.  To avoid the additional cost, we study methods for closing this gap by changing the sampling scheme, rather than the model.

To this end, our contributions are summarized as follows:
\begin{itemize}
\vspace{-\topsep}
  \setlength{\parskip}{0pt}
  \setlength{\itemsep}{0pt plus 1pt}
\item We propose an algorithm that enables universal guidance for diffusion models. 
Our proposed sampler evaluates the guidance models only on denoised images, rather than noisy latent states.  By doing so, we close the domain gap that has plagued standard guidance methods. 
%
This strategy provides the end-user with the flexibility to work with a wide range of guidance modalities and even multiple modalities simultaneously. 
The underlying diffusion model remains fixed and no fine-tuning of any kind is necessary. 
\item We demonstrate the effectiveness of our approach for a variety of different constraints such as \textit{classifier labels}, \textit{human identities}, \textit{segmentation maps}, \textit{annotations from object detectors}, and constraints arising from \textit{inverse linear problems}.
\end{itemize}

\section{Background}
We first briefly review the recent literature on the core framework behind diffusion models. Then, we define the problem setting of controlled image generation and discuss previous related 
works.

\subsection{Diffusion Models}

Diffusion models are strong generative models that proved powerful even when first introduced for image generation \citep{song2019generative, DDPM_Ho2020}.
The approach has been successfully extended to a number of domains, such as audio and text generation~\citep{kong2020diffwave,huang2022fastdiff,austin2021structure_diff,li2022diffusion_lm}.

We introduce (unconditional) diffusion formally, as it is helpful in describing the nuances of different types of models.  
A diffusion model is defined as a combination of a $T$-step forward process and a $T$-step reverse process. 
Conceptually, the forward process gradually adds Gaussian noise of different magnitudes to a clean data point $z_0$, while the reverse process attempts to gradually denoise a noisy input in hopes of recovering a clean data point.
More concretely,
given an array of scalars representing noise scales $\{\alpha_t\}_{t=1}^T$ and an initial, clean data point $z_0$, applying $t$ steps of the forward process to $z_0$ yields a noisy data point
\begin{equation}
z_t = \sqrt{\alpha_t} z_0 +  (\sqrt{1 - \alpha_t}) \epsilon ,\, \epsilon \sim \mathcal{N}(0, \mat I).
\end{equation}
A diffusion model is a learned denoising network $\epsilon_{\theta}$. 
It is trained so that for any pair $(z_0, t)$ and any sample of $\epsilon$,
\begin{equation}
\epsilon_{\theta}(z_t, t) \approx \epsilon = \frac{z_t - \sqrt{\alpha_t} z_0}{\sqrt{1 - \alpha_t}}.
\end{equation}


The reverse process takes the form $q(z_{t-1} | z_t, z_0)$ with various detail definitions,
where $q(\cdot|\cdot)$ is generally parameterized as a Gaussian distribution.
Different works also studied different approximations of the unknown $q(z_{t-1} | z_t, z_0)$ used to perform sampling.
For example, denoising diffusion implicit model (DDIM)~\citep{DDIM_Song2021} first computed a \emph{predicted} clean data point 
\begin{equation}\label{eq:pred_z0}
\hat z_0 = \frac{z_t - (\sqrt{1 - \alpha_t}) \epsilon_{\theta}(z_t, t)}{\sqrt{\alpha_t}},
\end{equation}
and sample $z_{t-1}$ from $q(z_{t-1} | z_t, \hat z_0)$ by replacing unknown $z_0$ with $\hat z_0.$
On the other hand, while the details of individual sampling methods vary, all sampling methods produce $z_{t-1}$ based on current sample $z_t$, current time step $t$ and a predicted noise $\hat\epsilon.$ To ease the notation burden, we define a function $S(\cdot,\cdot,\cdot)$ as an abstraction of the sampling method, where $z_{t-1} = S(z_t, \hat\epsilon, t).$


\subsection{Controlled Image Generation}

In this paper, we focus on controlled image generation with various constraints.
Consider a differentiable guidance function $f$, for example a CLIP feature extractor or a segmentation network.  When applied to an image, we obtain a vector $c = f(x).$ We also consider a function $\ell(\cdot, \cdot)$ that measures the closeness of two vectors $c$ and $c'$. Given a particular choice of $c,$ which we call a {\em prompt}, the corresponding constraint (based on $c, \ell,$ and $f$) is formalized as $\ell(c, f(z)) \approx 0,$ and we aim to generate a 
 sample $z$ from the image distribution satisfying the constraint.  In plain words, we want to generate an in-distribution image that matches the prompt.

Prior work that studied controlled generative diffusion mainly falls into two categories. 
We refer to the first category as conditional image generation, and the second category as guided image generation. Next, we discuss the characteristics of each category and better situate our work among existing methods.

\paragraph{Conditional Image Generation.}

Methods from this category require training new diffusion models that accept the prompt as an additional input~\citep{ho2022classifierfree,bansal2022colddiff,nichol2021glide,whang2022debluriter,wang2022segmentation_diffusion}.
For example, \citet{ho2022classifierfree} proposed classifier-free guidance using class labels as prompts, and trained a diffusion model by linear interpolation between unconditional and conditional outputs of the denoising networks. 
\citet{bansal2022colddiff} studied the case where the guidance function is a known linear degradation operator, and trained a conditional model to solve linear inverse problems.
\citet{nichol2021glide} further extended classifier-free guidance to text-conditional image generation with descriptive phrases as prompts, and trained a diffusion model to enforce the similarity between the CLIP~\citep{radford2021clip} representations of the generated images and the text prompts. These methods are successful across different types of constraints, however the requirement to retrain the diffusion model makes them computationally intensive. 

\paragraph{Guided Image Generation.}
Works in this category employed a frozen pre-trained diffusion model as a foundation model, but modify the sampling method to guide the image generation with feedback from the guidance function. Our method falls into this category. Prior work that studied guided image generation did so with a variety of restrictions and external guidance functions~\citep{dhariwal21diffusion_beats_gan,bahjat2022ddrm,wang2022zeroshot_diffusion,chung2022general_inverse,lugmayr2022inpaint_diffusion,chung2022maniold_diffusion,graikos2022diffusion_plug_and_play}. For example, \citet{dhariwal21diffusion_beats_gan} proposed classifier guidance, where they trained a classifier on images of different noise scales as the guidance function $f$, and included gradients of the classifier during the sampling process. However, a classifier for noisy images is domain-specific and generally not readily available -- an issue our method circumvents. \citet{wang2022zeroshot_diffusion} assumed the external guidance functions to be linear operators, and generated the component of images residing in the null space of linear operators with the foundation model. Unfortunately, extending that method to handle non-linear guidance functions is non-trivial. \citet{chung2022general_inverse} studied general guidance functions, and modified the sampling process with the gradient of guidance function calculated on the expected denoised images. Nevertheless, the authors only presented results with simpler non-linear guidance functions such as non-linear blurring.

In this work, we study universal guidance algorithms for guided image generation with diffusion models using any off-the-shelf guidance functions $f$, such as object detection or segmentation networks. 

\section{Universal Guidance}
\label{sec:universal-guidance}

We propose a guidance algorithm that augments the image sampling method of a diffusion model to include guidance from an off-the-shelf auxiliary network. Our algorithm is motivated by an empirical observation that the reconstructed clean image $\hat z_0$ obtained by \cref{eq:pred_z0}, while naturally imperfect, is still appropriate for a generic guidance function to provide informative feedback to guide the image generation. 
In \cref{sec:universal_guidance:g1}, we motivate our \emph{forward universal guidance} by extending classifier guidance~\cite{dhariwal21diffusion_beats_gan} to leverage this observation and handle generic guidance functions. In \cref{sec:universal_guidance:g2}, we propose a supplementary \emph{backward universal guidance} to help enforce the generated image to satisfy the constraint based on the guidance function $f$. In \cref{sec:universal_guidance:repeat}, we discuss a simple yet helpful self-recurrence trick to empirically improve the fidelity of generated images.


\subsection{Forward Universal Guidance}
\label{sec:universal_guidance:g1}
To guide the generation with information from the external guidance function $f$ and the loss function $\ell$, an immediate thought is to extend classifier guidance~\citep{dhariwal21diffusion_beats_gan} to accept any general guidance function. Concretely, given a class prompt $c$, classifier guidance performs classification-guided sampling by replacing $\epsilon_{\theta} (z_t, t)$ in each sampling step $S(z_t, t)$ with 
\begin{equation}\label{eq:classifier_guidance_original}
    \hat \epsilon_{\theta} (z_t, t) = \epsilon_{\theta} (z_t, t) - \sqrt{1 - \alpha_t} \grad{z_t} \log p(c | z_t).
\end{equation}
Defining $\ell_{ce}(\cdot,\cdot)$ to be the cross-entropy loss and $f_{cl}$ to be the guidance function that outputs classification probability, \cref{eq:classifier_guidance_original} can be re-writtern as
\begin{equation}\label{eq:classifier_guidance}
    \hat \epsilon_{\theta} (z_t, t) = \epsilon_{\theta} (z_t, t) + \sqrt{1 - \alpha_t} \grad{z_t} \ell_{ce} (c, f_{cl}(z_t)).
\end{equation}
However, directly replacing $f_{cl}$ and $\ell_{ce}$ with any off-the-shelf guidance and loss functions does not work in practice, as $f$ is most likely trained on clean images and fails to provide meaningful guidance when the input is noisy.

To address the issue, we leverage the fact that $\epsilon_{\theta}(z_t, t)$ predicts the noise added to the data point, and we can therefore obtain a \emph{predicted} clean image $\hat z_0$ by \cref{eq:pred_z0}. We propose to instead calculate the guidance based on the predicted clean data point as
\begin{equation} \label{eq:guidance_1}
     \hat \epsilon_{\theta} (z_t, t)  = \epsilon_{\theta} (z_t, t) + s(t) \cdot \grad{z_t} \ell (c, f(\hat z_0))
\end{equation}
where $s(t)$ controls the guidance strength for each sampling step and
\[
\grad{z_t} \ell (c, f(\hat z_0)) = \grad{z_t} \ell \left(c, f\left(\frac{z_t - \sqrt{1 - \alpha_t} \epsilon_{\theta}(z_t, t)}{\sqrt{\alpha_t}}\right)\right)
\]
as in \cref{eq:pred_z0}.
We term \cref{eq:guidance_1} forward universal guidance, or forward guidance in short. In practice, applying forward guidance effectively brings the generated image closer to the prompt while keeping the generation trajectory in the data manifold. We note that a related approach is also studied in~\cite{chung2022general_inverse}, where the guidance step is computed based on $E[z_0|z_t]$. The approach drew inspiration from the score-based generative framework~\cite{scoreSDE_2021Song}, but resulted in a different update method.



\begin{algorithm}[t]
   \caption{Universal Guidance}
   \label{alg:ug}
    \begin{algorithmic}
    \STATE {\bfseries Parameter:} Recurrent steps $k$, gradient steps $m$ for backward guidance and guidance strength $s(t)$,
    \STATE {\bfseries Required:} $z_T$ sampled from $\mathcal{N}(0, I)$, diffusion model $\epsilon_{\theta}$, noise scales $\{\alpha_t\}_{t=1}^T$, guidance function $f$, loss function $\ell$, and prompt $c$
    \FOR{\textit{t} $=T, T-1,\dots,1$}
    \FOR{\textit{n} $=1, 2, \dots, k$}
    \STATE Calculate $\hat z_0$ as \cref{eq:pred_z0}
    \STATE Calculate $\hat \epsilon_\theta$ using forward universal guidance as \cref{eq:guidance_1}
    \IF {$m > 0$}
        \STATE Calculate $\Delta z_0$ by minimizing \cref{eq:g2_optim} with $m$ steps of gradient descent
        \STATE Perform backward universal guidance by \\ $\hat \epsilon_\theta \gets \hat\epsilon_\theta - \sqrt{\alpha_t/(1 - \alpha_t)} \Delta z_0$ ( \cref{eq:g2_guided_eps}) 
    \ENDIF
    \STATE $z_{t-1} \gets S(z_t, \hat \epsilon_\theta, t)$ 
    \STATE $\epsilon' \sim  \mathcal{N}(0, I)$ 
    \STATE $z_t \gets \sqrt{\alpha_t/\alpha_{t-1}} z_{t-1} + \sqrt{1 - \alpha_t/\alpha_{t-1}} \epsilon'$ 
    \ENDFOR
    \ENDFOR
    \end{algorithmic}
\end{algorithm}

\subsection{Backward Universal Guidance}
\label{sec:universal_guidance:g2}
As will be shown in \cref{sec:exp:imagenet}, we observe that forward guidance sometimes over-prioritizes maintaining the ``realness'' of the image, resulting in an unsatisfactory match with the given prompt. Simply increasing the guidance strength $s(t)$ is suboptimal, as this often results in instability as the image moves off the manifold faster than the denoiser can correct it.

To address the issue, we propose backward universal guidance, or backward guidance in short, to supplement forward guidance and help enforce the generated image to satisfy the constraint.
The key idea of backward guidance is to optimize for a clean image that best matches the prompt based on $\hat z_0$, and linearly translate the guided change back to the noisy image space at step $t$.
Concretely, instead of directly calculating $\grad{z_t} \ell(c, f(\hat z_0))$, we compute a guided change $\Delta z_0$ in clean data space as
\begin{equation}\label{eq:g2_optim}
    \Delta z_0  =  \arg\min_{\Delta} \ell(c, f(\hat z_0+\Delta)).
\end{equation} 
Empirically, we solve \cref{eq:g2_optim} with $m$-step gradient descent, where we use $\Delta = 0$ as a starting point.
Since $\hat z_0 + \Delta z_0$ minimizes $\ell(c, f(z))$ directly, $\Delta z_0$ is the change in clean data space that best enforces the constraint. Then, we translate $\Delta z_0$ back to the noisy data space of $z_t$ by calculating the \emph{guided denoising prediction} $\tilde \epsilon$ that satisfies
\begin{equation}\label{eq:g2_guided_eps_def}
z_t = \sqrt{\alpha_t} (\hat z_0 + \Delta z_0) + \sqrt{1 - \alpha_t} \tilde \epsilon.
\end{equation}
Reusing \cref{eq:pred_z0}, we can rewrite $\tilde \epsilon$ as an augmentation to the original denoising prediction $\epsilon_{\theta} (z_t, t)$ by
\begin{equation}\label{eq:g2_guided_eps}
    \tilde \epsilon = \epsilon_{\theta}(z_t, t) - \sqrt{\alpha_t/(1 - \alpha_t)} \Delta z_0.
\end{equation}

Comparing to forward guidance, backward guidance (as \cref{eq:g2_guided_eps}) produces an optimized direction for the generated image to match the given prompt, and hence prioritizes enforcing the constraint. Furthermore, calculation of a gradient step for \cref{eq:g2_optim} is computationally cheaper than forward guidance (\cref{eq:guidance_1}), and we can therefore afford to solve \cref{eq:g2_optim} with multiple gradient steps, further improving the match with the given prompt. \\

We note that the names ``forward'' and ``backward'' are used analogously to the forward and backward Euler methods.

\subsection{Per-step Self-recurrence}
\label{sec:universal_guidance:repeat}

Unfortunately, when we apply our universal guidance to standard generation pipelines, we often find images with artifacts and strange behaviors that clearly separate them from natural images. Similar observations have been made in~\cite{lugmayr2022inpaint_diffusion,wang2022zeroshot_diffusion}, where linear guidance functions are studied. Our attempts to prioritize realness by decreasing $s(t)$ proved ineffective; the sweet spot that both ensures the realness and guidance constraint satisfaction doesn't always exist, especially for complex guidance functions. We conjecture that the guidance direction produced by our universal method is not always related to the realness of the images when the guidance function creates too much information loss, causing the image to stray from the natural image sampling trajectory.

\begin{figure}[t]
    \centering
    \includegraphics[width=\columnwidth]{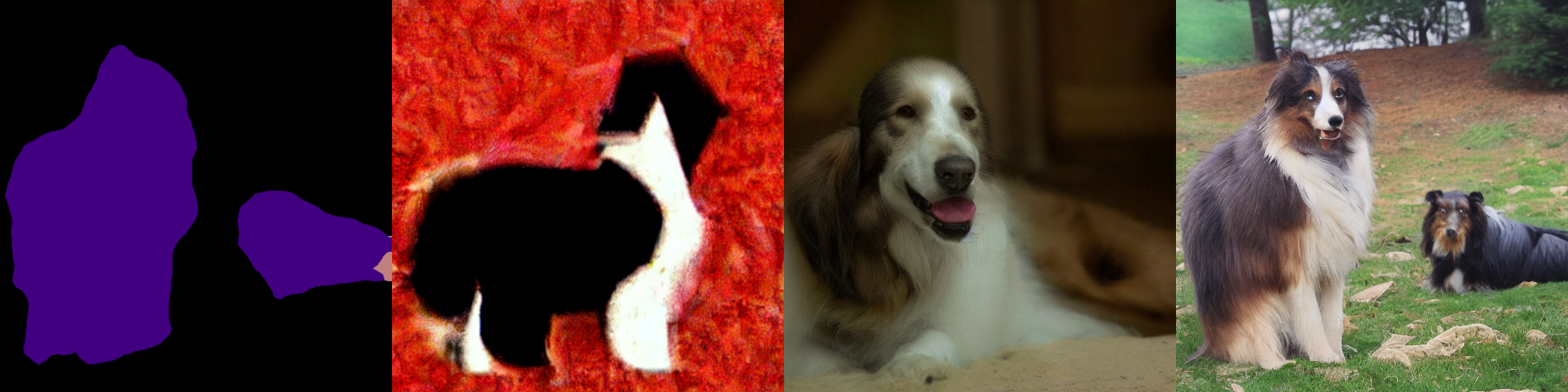}
    \caption{An example of how self-recurrence helps segmentation-guided generation. The left-most figure is the given segmentation map, and the images generated with recurrence steps of 1, 4 and 10 follow in order.}
    \label{fig:recur}
    \vspace{-.3cm}
\end{figure}

Inspired by~\cite{lugmayr2022inpaint_diffusion,wang2022zeroshot_diffusion}, we address the issue by applying per-step self-recurrence. More concretely, after $z_{t-1} = S(z_t, \hat\epsilon_t, t)$ is sampled, we re-inject random Gaussian noise $\epsilon'\sim \mathcal{N}(0, \mat I)$ to $z_{t-1}$ to obtain $z'_t$ by
\begin{equation}\label{eq:self_recurrence}
    z'_t = \sqrt{\alpha_t/\alpha_{t-1}} \cdot z_{t-1} + \sqrt{1- \alpha_t/\alpha_{t-1}} \cdot \epsilon'.
\end{equation}
\cref{eq:self_recurrence} ensures $z'_t$ to have proper noise scale for input at time step $t$. We repeat the self-recurrence $k$ times before continuing the sampling for step $t-1$. Intuitively, the self-recurrence allows exploration of different regions of the data manifold at the same noise scale, allowing more budget to find a solution that satisfies both guidance and image quality. 
Empirically, we find that our self-recurrence can keep the realness of the generated image with a proper guidance strength $s(t)$ that ensures the match with the given prompt. We illustrate an example of how self-recurrence improves the harmony of generated images in \cref{fig:recur}.

We summarize our universal guidance algorithm composed of forward universal guidance, backward universal guidance and per-step self-recurrence in \cref{alg:ug}. For simplicity, the algorithm assumes only one guidance function, but can be easily adapted to handle multiple pair of $(f, l)$. Additionally, the objectives of the forward and backward guidance do not have to be identical, allowing different ways to simultaneously utilize multiple guidance functions. 

 \newcommand\gw{\columnwidth}
 \newcolumntype{i}{@{\hspace{0ex}}
   >{\collectcell\includepic}c<{\endcollectcell}}
 \newcommand\nextline[1]{\hbox{\strut {\fontsize{33}{40}\selectfont \emph{#1}}}}

\section{Experiments}
In this section, we present results testing our proposed universal guidance algorithm against a wide variety of guidance functions. Specifically, we experiment with \href{https://stablediffusionweb.com/}{Stable Diffusion}~\cite{rombach2022stablediff}, a diffusion model that is able to perform text-conditional generation by accepting text prompt as additional input, and experiment with a purely unconditional diffusion model trained on ImageNet~\cite{deng2009imagenet}, where we use \href{https://github.com/openai/guided-diffusion}{pre-trained model} provided by OpenAI~\cite{dhariwal21diffusion_beats_gan}. We note that Stable Diffusion, while being a text-conditional generative model, can also perform unconditional image generation by simply using an empty string for the text prompt. We first present the experiment on Stable Diffusion for different guidance functions in \cref{sec:exp:stable_diff}, and present the results on ImageNet diffusion model in \cref{sec:exp:imagenet}.



\begin{table}[t]
\centering
\resizebox{\columnwidth}{!}{%
\begin{tabular}{cc}
\toprule
    \fontsize{18}{22}\selectfont Conditional Stable-Diffusion & \fontsize{16}{22}\selectfont Guided Stable-Diffusion \\ 
    \midrule  \\

    \multicolumn{2}{c}{\fontsize{20}{22}\selectfont \emph{A photograph of an astronaut riding a horse.}} \\

    \includegraphics[width=\gw,valign=m]{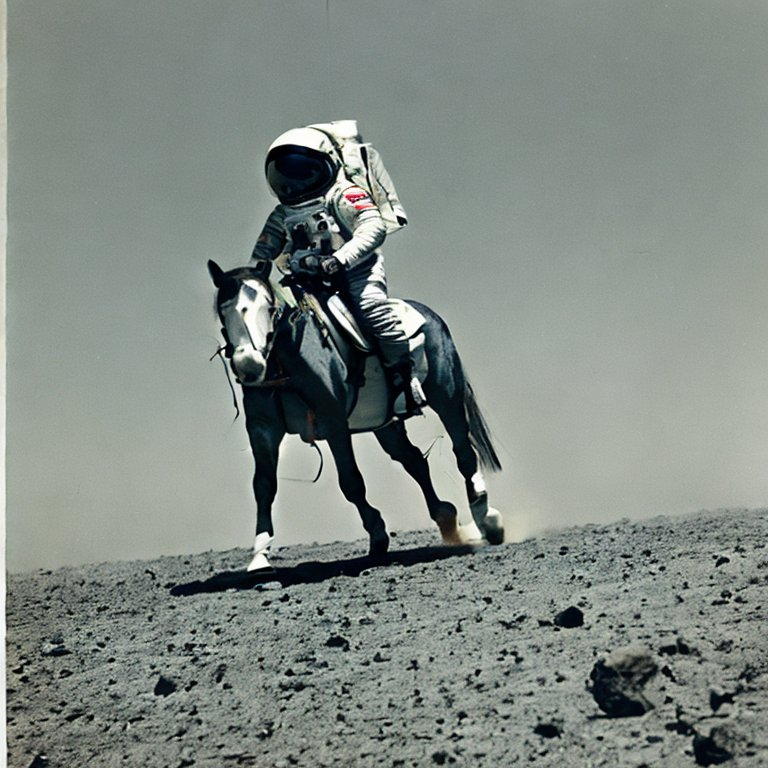} &
    \includegraphics[width=\gw,valign=m]{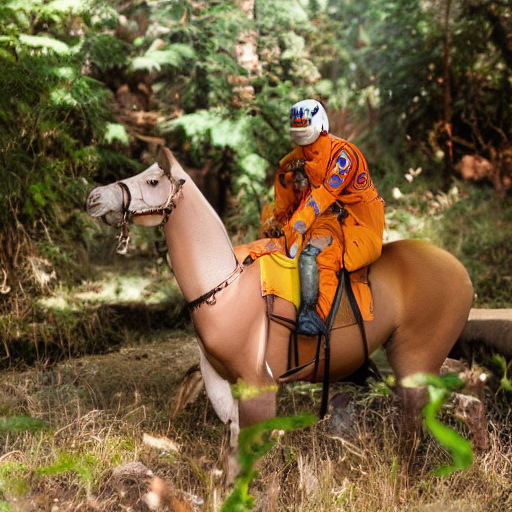} \\
    \\

    \multicolumn{2}{c}{\fontsize{20}{22}\selectfont \emph{An oil painting of a corgi wearing a party hat.}} \\
    
    \includegraphics[width=\gw,valign=m]{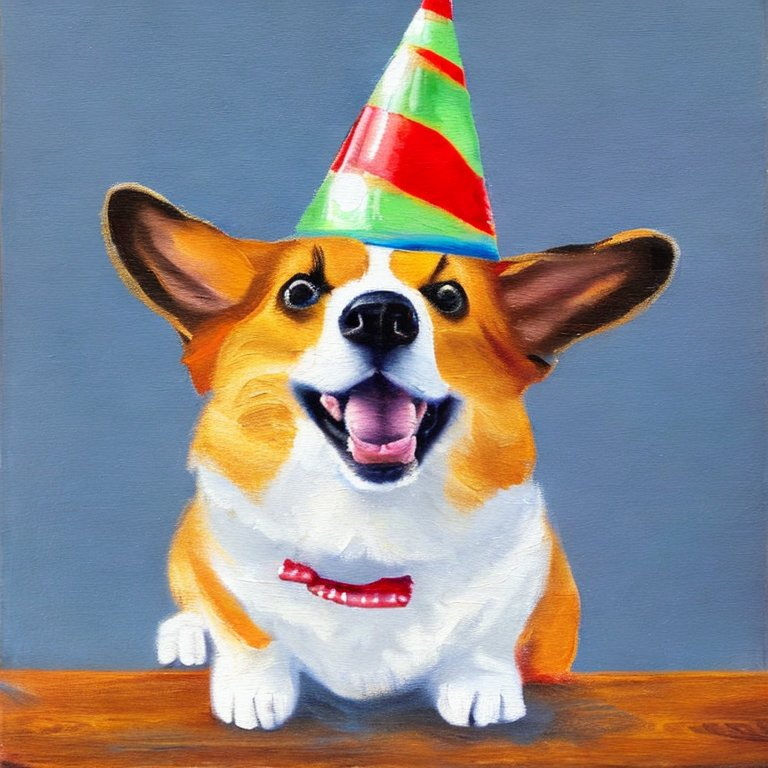} &
    \includegraphics[width=\gw,valign=m]{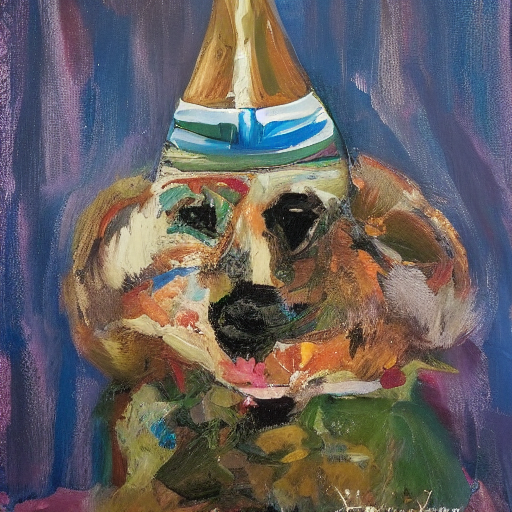} \\
     \\
\end{tabular}
}
\captionof{figure}[]{We compare the ability to match given text prompts between our universal guidance algorithm and text-conditional model trained from scratch. The results demonstrate that our universal algorithm is comparable to specialized conditional model on the ability to generate quality images that satisfy the text constraints.
}
\label{fig:clip_sd}
\vspace{-.2cm}
\end{table}

\begin{table}[t]
\vspace{18pt}
\centering
\resizebox{\columnwidth}{!}{%
\begin{tabular}{cccc}
    {\fontsize{33}{40}\selectfont \backslashbox{Guide}{Prompt}} &
    
    \vtop{\vspace{-50pt}\hsize=\columnwidth \hangindent=0em \nextline{Walker hound,} \nextline{Walker foxhound }\nextline{under water.}}&
    
    \vtop{\vspace{-50pt}\nextline{Walker hound,}\nextline{Walker foxhound }\nextline{on snow.}}&
    
    \vtop{\vspace{-50pt}\nextline{Walker hound,}\nextline{Walker foxhound}\nextline{as an oil painting.}}\\
    \\
    \vtop{\nextline{(N/A)}} &
    \includegraphics[width=\gw,valign=m]{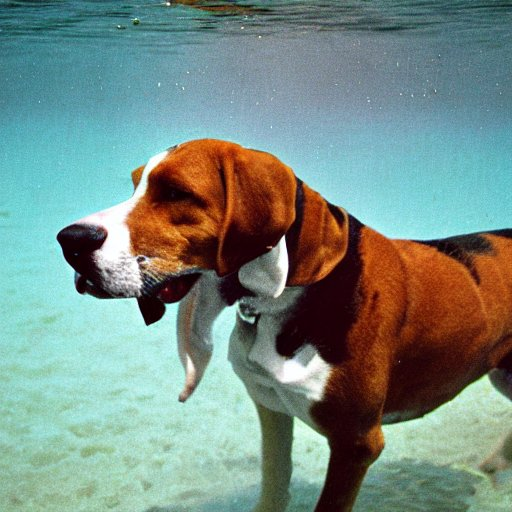} &
    \includegraphics[width=\gw,valign=m]{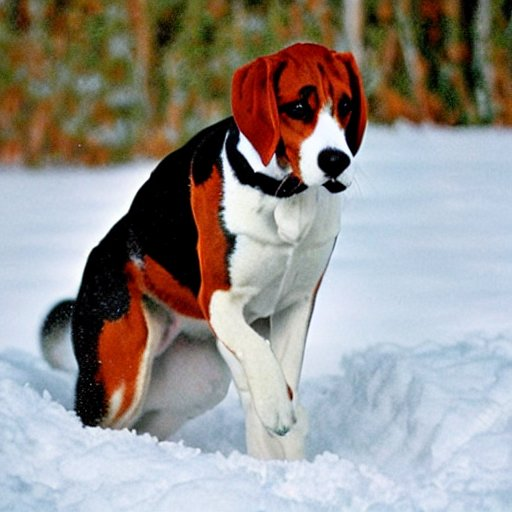} &
    \includegraphics[width=\gw,valign=m]{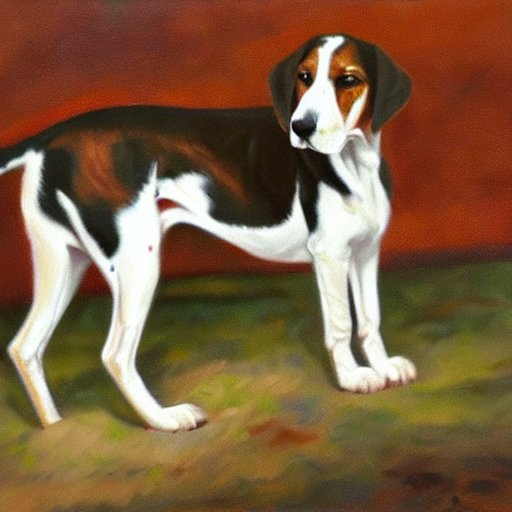}  \\

    \includegraphics[width=\gw,valign=m]{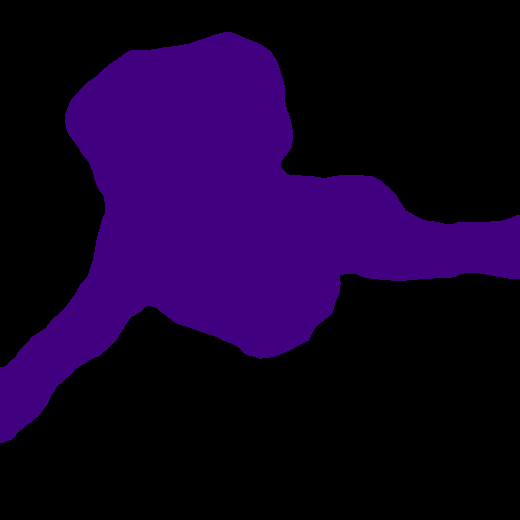} &
    \includegraphics[width=\gw,valign=m]{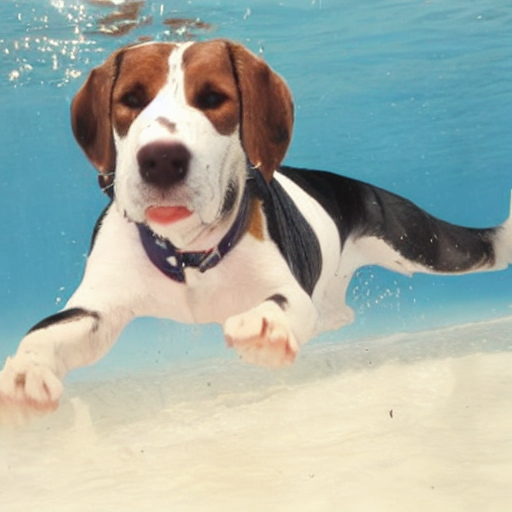} &
    \includegraphics[width=\gw,valign=m]{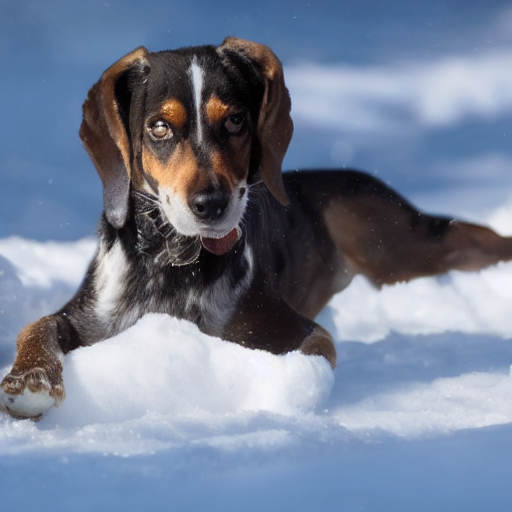} &
    \includegraphics[width=\gw,valign=m]{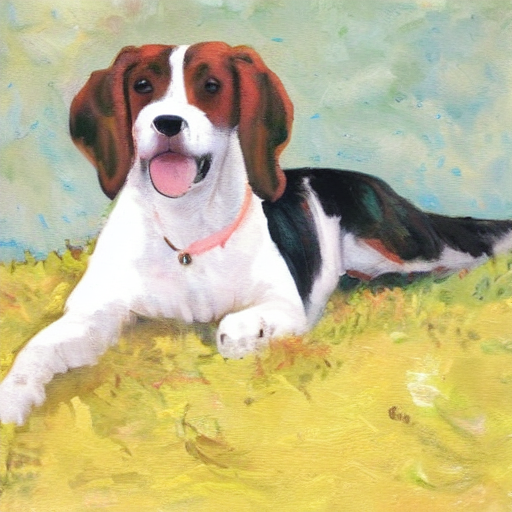} \\
    
    \includegraphics[width=\gw,valign=m]{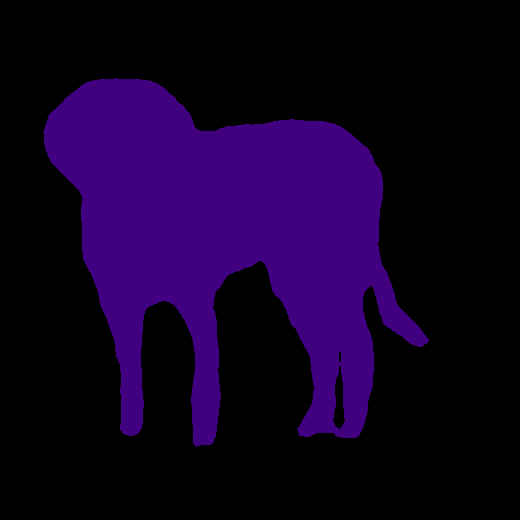} &
    \includegraphics[width=\gw,valign=m]{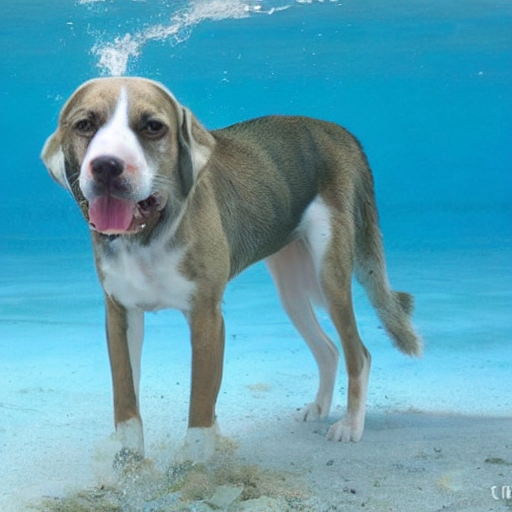} &
    \includegraphics[width=\gw,valign=m]{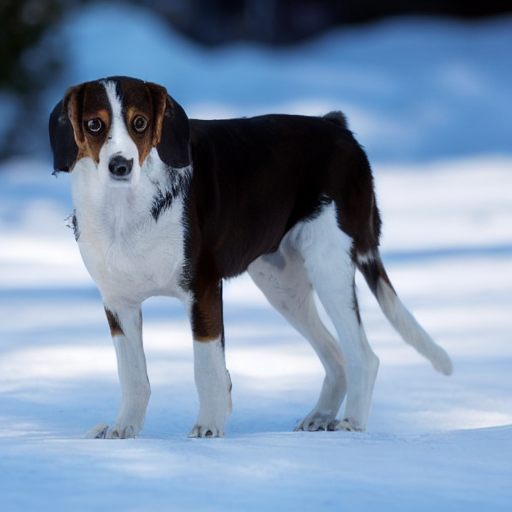} &
    \includegraphics[width=\gw,valign=m]{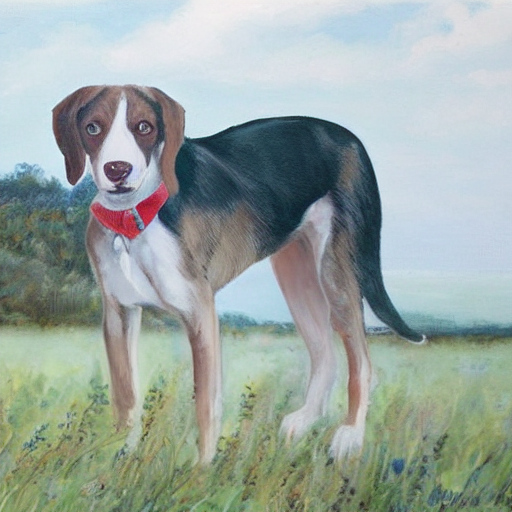} \\

    \includegraphics[width=\gw,valign=m]{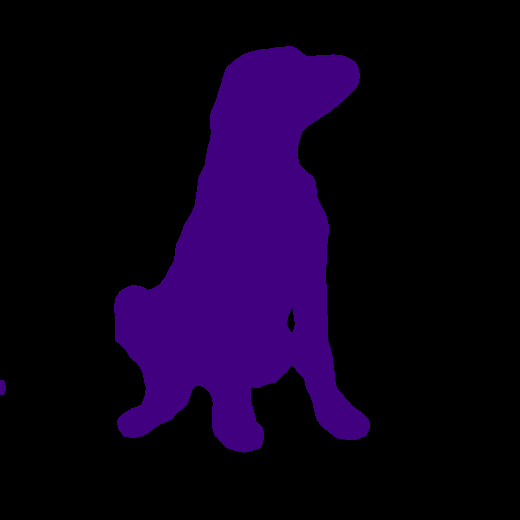} &
    \includegraphics[width=\gw,valign=m]{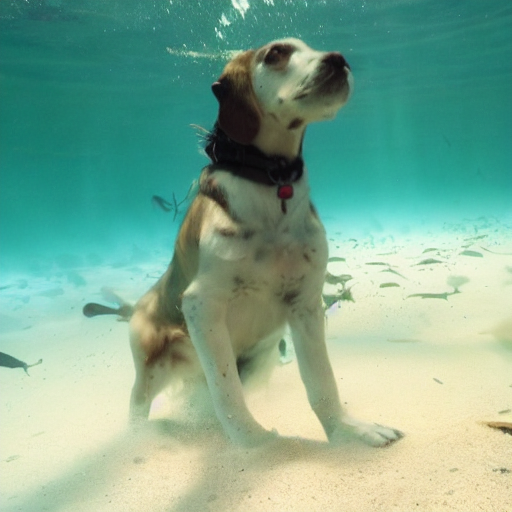} &
    \includegraphics[width=\gw,valign=m]{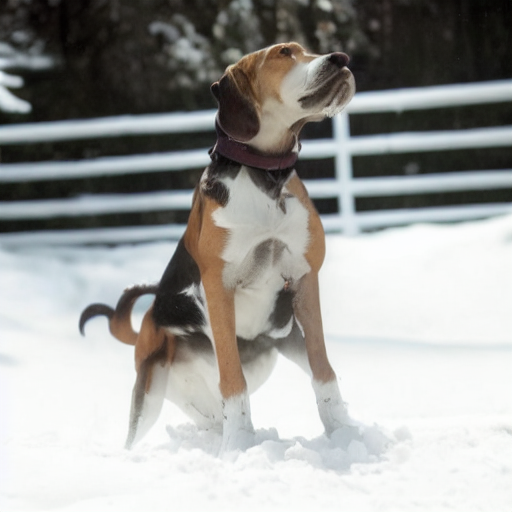} &
    \includegraphics[width=\gw,valign=m]{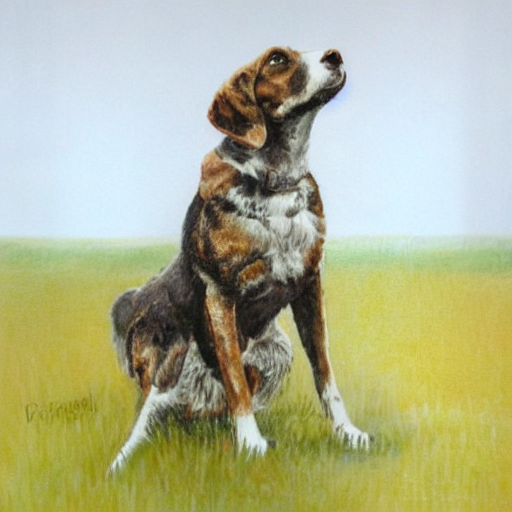} \\
    
\end{tabular}}
\captionof{figure}[]{In addition to matching the text prompts (above each column), these images are guided by an image segmentation pipeline. Each column contains examples of images generated to match the prompt and the segmentation map in the left-most column. The top-most row contains examples generated without guidance.}
\label{fig:segmentation_sd}
\vspace{-.2cm}
\end{table}

\begin{table}[ht]
\vspace{18pt}
\centering
\resizebox{\columnwidth}{!}{%
\begin{tabular}{cccc}
    {\fontsize{33}{40}\selectfont \backslashbox{Guide}{Prompt}} &
    \vtop{\vspace{-100pt}\hsize=\columnwidth \hangindent=0em \nextline{Headshot of a} \nextline{person with }\nextline{blonde hair}\nextline{with space}\nextline{background.}}&
    \vtop{\vspace{-50pt}\nextline{Headshot of a}\nextline{woman made}\nextline{of marble.}}&
    \vtop{\vspace{-50pt}\nextline{A headshot of}\nextline{a woman looking}\nextline{like Lara Croft.}}\\
    \\
    \vtop{\nextline{(N/A)}} &
    \includegraphics[width=\gw,valign=m]{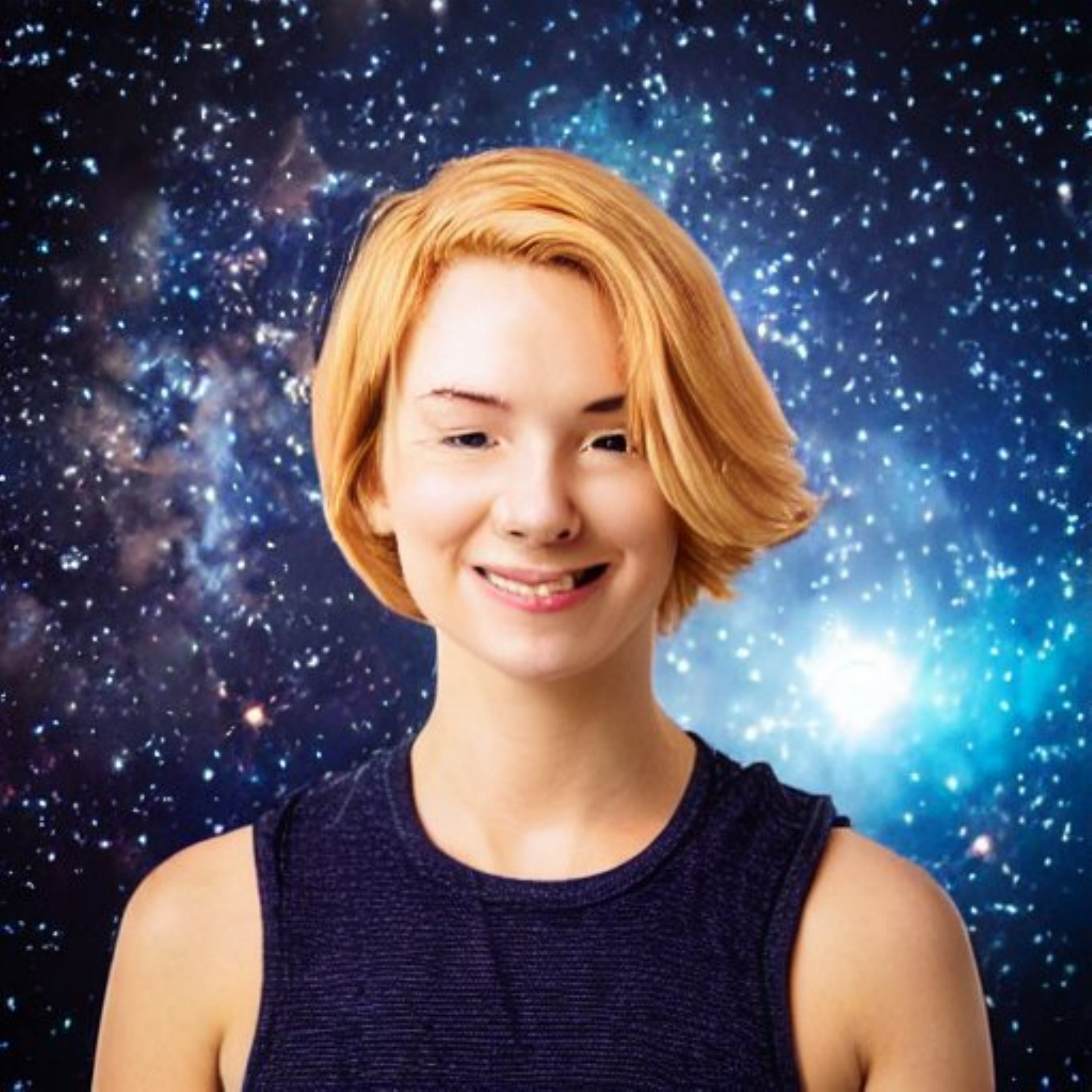} &
    \includegraphics[width=\gw,valign=m]{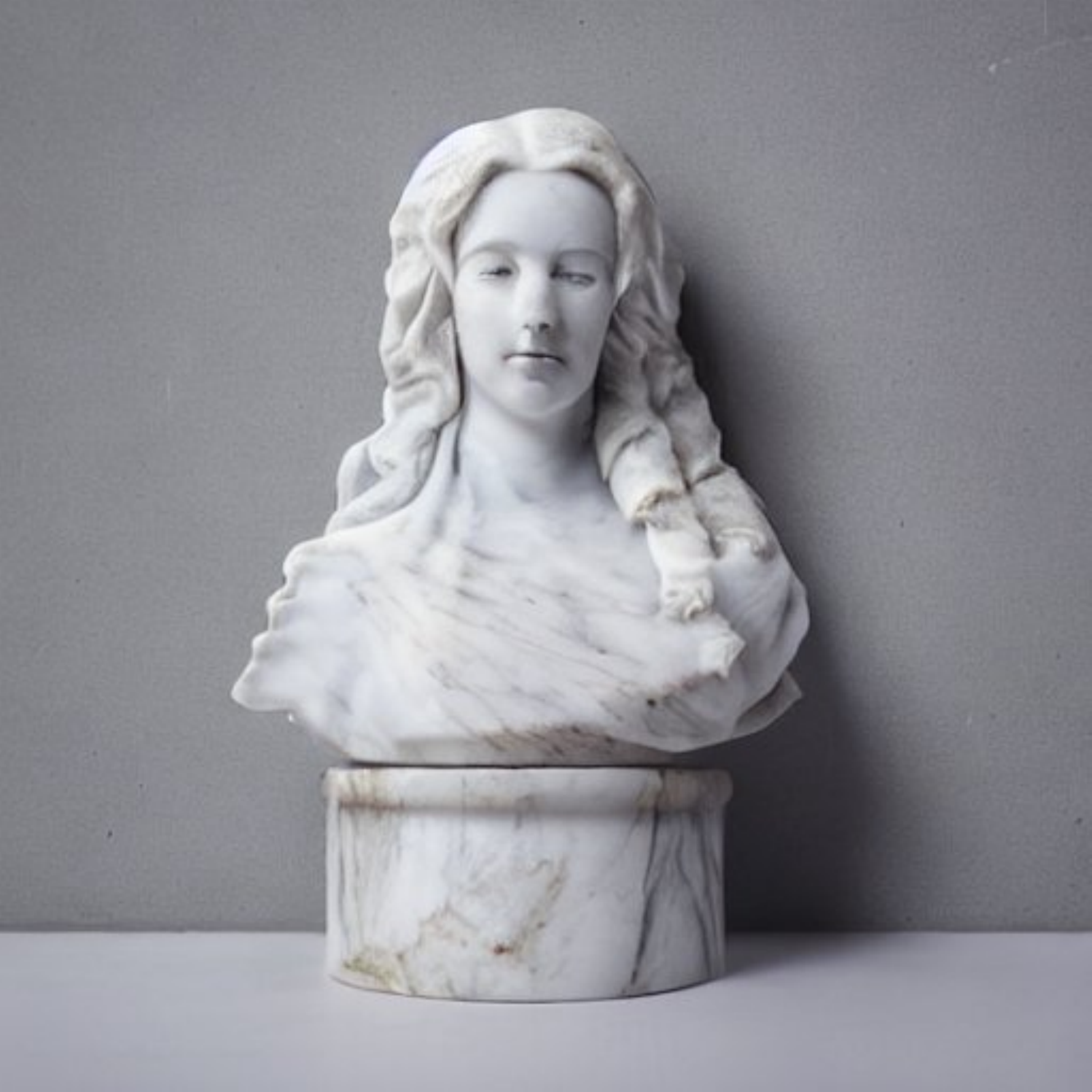} &
    \includegraphics[width=\gw,valign=m]{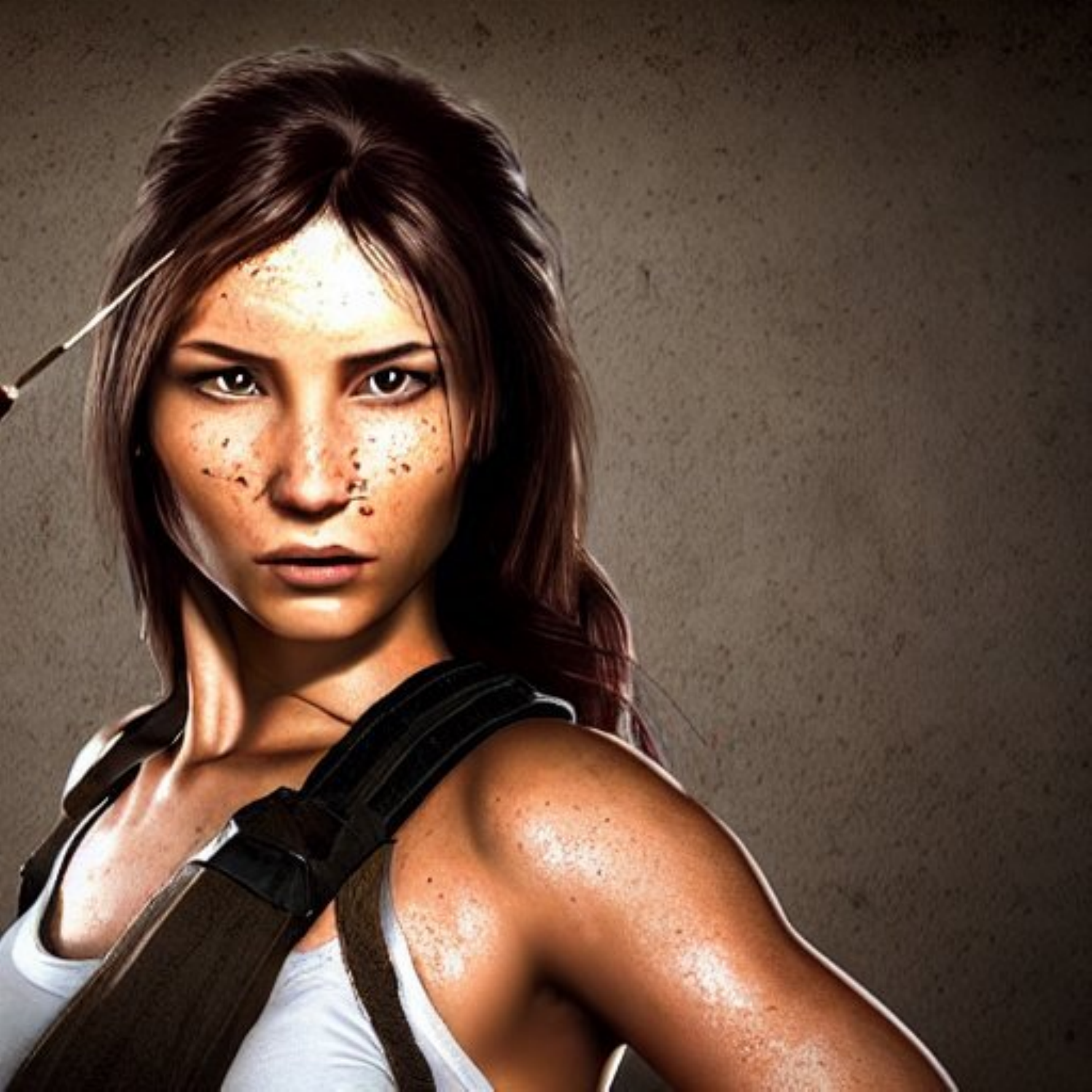}  \\

    \includegraphics[width=\gw,valign=m]{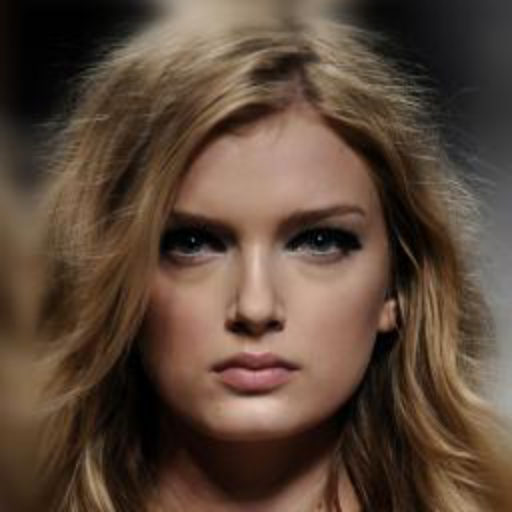} &
    \includegraphics[width=\gw,valign=m]{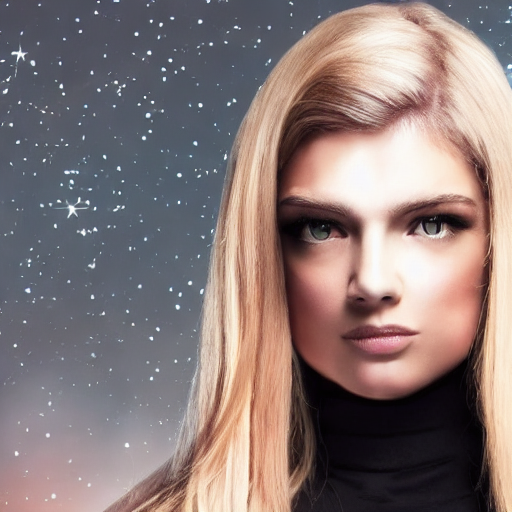} &
    \includegraphics[width=\gw,valign=m]{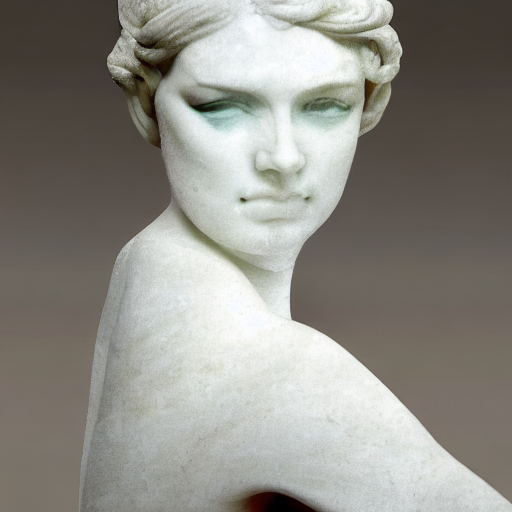} &
    \includegraphics[width=\gw,valign=m]{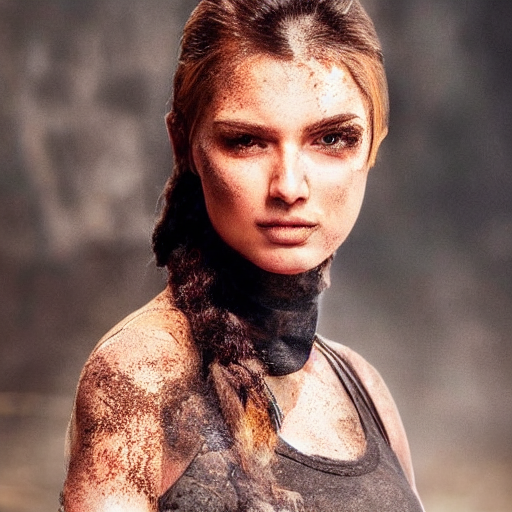} \\
    
    \includegraphics[width=\gw,valign=m]{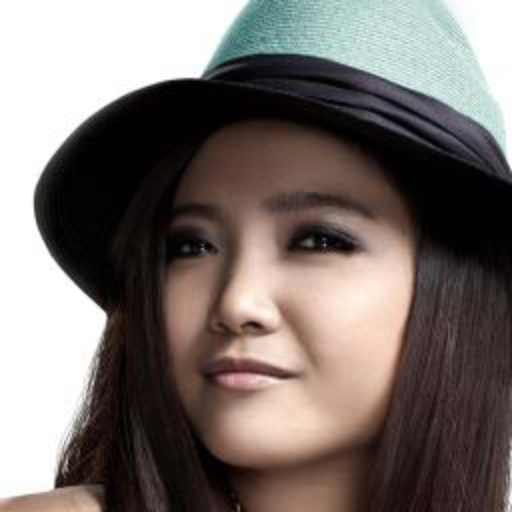} &  
    \includegraphics[width=\gw,valign=m]{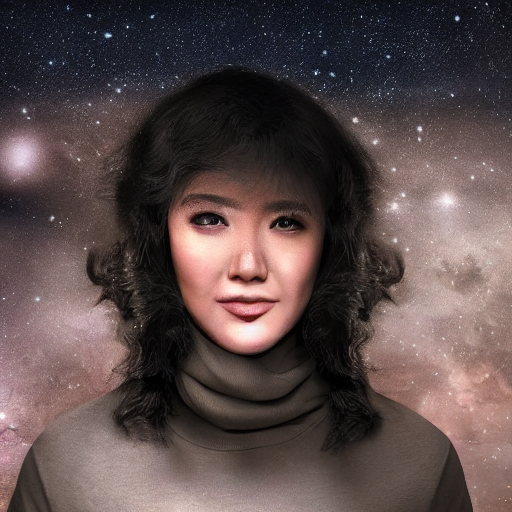} &
    \includegraphics[width=\gw,valign=m]{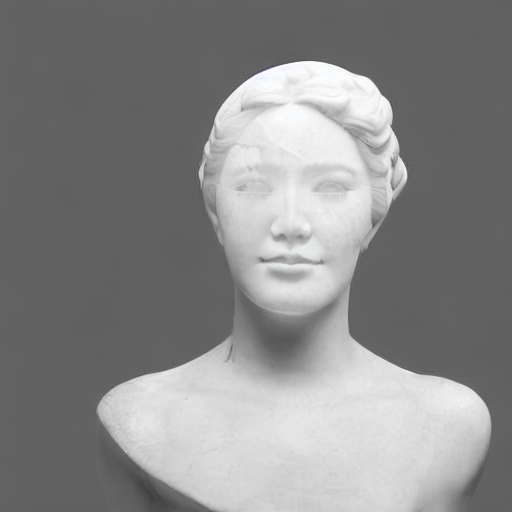} &
    \includegraphics[width=\gw,valign=m]{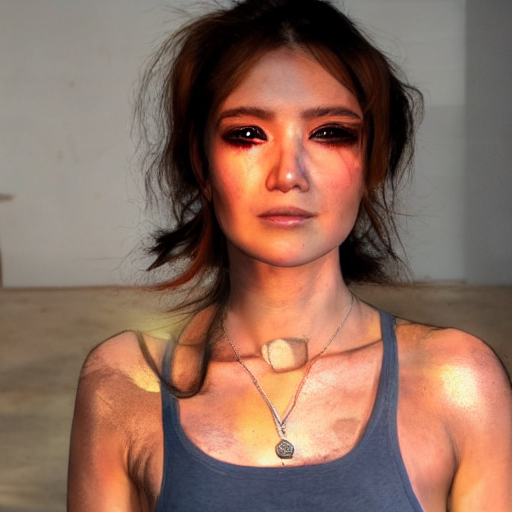} \\

    \includegraphics[width=\gw,valign=m]{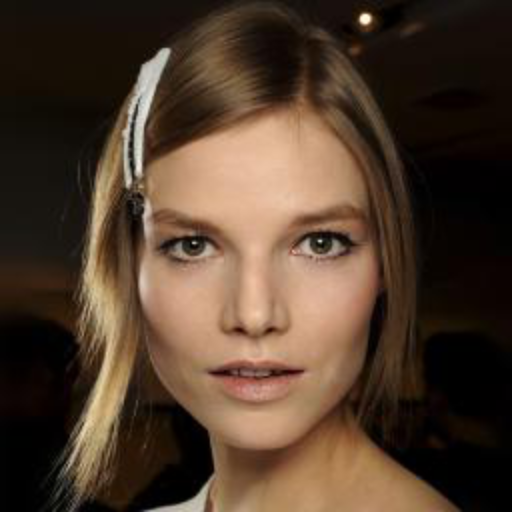} &  
    \includegraphics[width=\gw,valign=m]{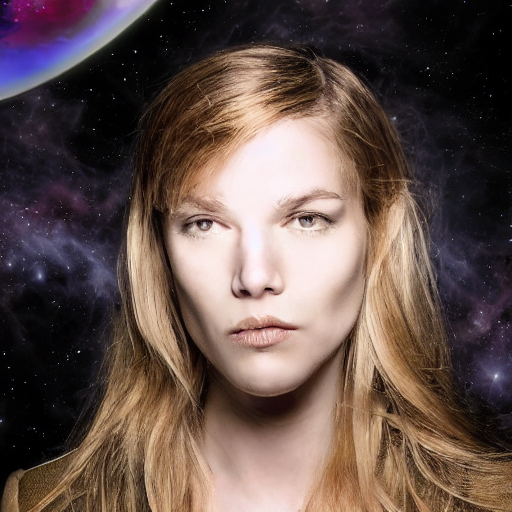} &
    \includegraphics[width=\gw,valign=m]{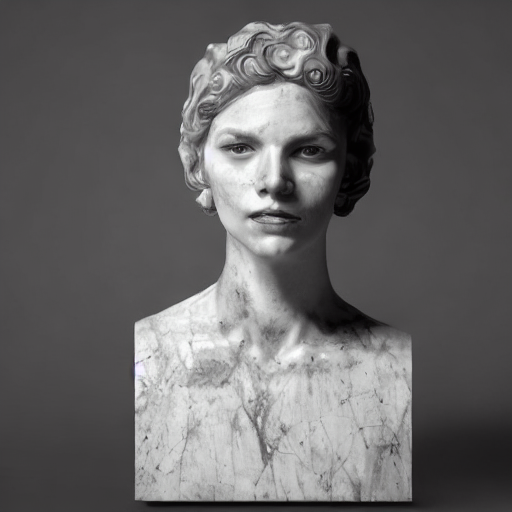} &
    \includegraphics[width=\gw,valign=m]{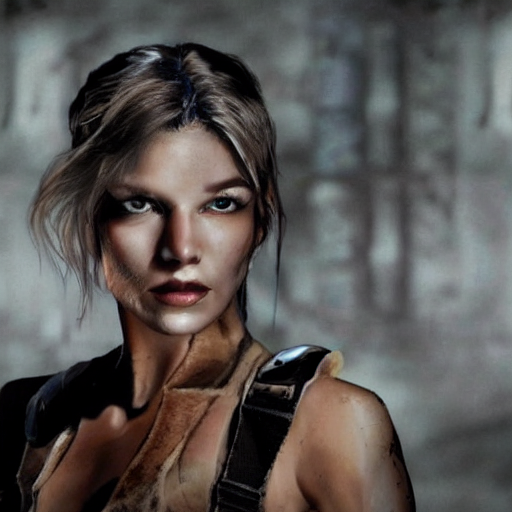} \\
    
\end{tabular}}
\captionof{figure}[]{In addition to matching the text prompts (above each column), these images are guided by a facial recognition system. Each column contains examples of images generated to match the prompt and the identity of the images in the left-most column. The top-most row contains examples generated without guidance.}
\label{fig:face_sd}
\vspace{-.2cm}
\end{table}

\subsection{Results for Stable Diffusion}
\label{sec:exp:stable_diff}
In this section, we present the results of guided image generation using Stable Diffusion as the foundation model. The guidance functions we experiment with include the CLIP feature extractcor~\cite{radford2021clip}, a segmentation network, a face recognition network and an object detection network. For experiments on Stable Diffusion, we discover that applying forward guidance already produce high-quality images that match the given prompt, and hence set $m=0$. To perform forward guidance on Stable Diffusion, we forward the predicted clean latent variable computed by \cref{eq:pred_z0} through the image decoder of Stable Diffusion to obtain predicted clean images. We discuss the results and implementation details for each guidance function in its corresponding subsection.

\paragraph{CLIP Guidance.}
CLIP~\cite{radford2021clip} is a state-of-the-art text-to-image similarity model developed by OpenAI.
To apply our algorithm to text-guided image generation, we use the image feature extractor of CLIP as the guidance function. We construct a loss function that calculates the negative cosine similarity between an image embedding and the CLIP text embedding produced by a given text prompt. We use $s(t) = 10 \sqrt{1-\alpha_t}$ and $k=8$ and use Stable Diffusion as an unconditional image generator.

\looseness -1 We generate images guided by a number of text prompts.
To further assess our universal guidance algorithm and compare guidance and conditioning, we also generate images using classical, text-conditional generation by Stable Diffusion with identical prompts as inputs, and summarize the results in \cref{fig:clip_sd}. 
The results in \cref{fig:clip_sd} show that our algorithm can guide the generation to produce high-quality images that match the given text description, and are comparable with images generated by the specialized text-conditioning model.



\paragraph{Segmentation Map Guidance.}
To perform guided image generation using a segmentation map as prompt, we use a MobileNetV3-Large~\cite{howard2019mobilenet_v3} with a segmentation head, and a \href{https://pytorch.org/vision/main/models/generated/torchvision.models.segmentation.lraspp_mobilenet_v3_large.html}{publicly available} pre-trained model in PyTorch~\cite{paszke2019pytorch}. As the segmentation network outputs per-pixel classification probability, we construct a loss function $\ell$ as the sum of per-pixel cross-entropy loss between a given prompt and the predicted segmentation of generated images.  We set $s(t) = 400 \cdot \sqrt{1 - \alpha_t}$ and  $k=10$.

In our experiment, we combine segmentation maps that depict objects of different shapes with new text prompts. We use the text prompt as a fixed additional input to Stable Diffusion to perform text-conditional sampling, and guide the text-conditional generated images to match the given segmentation maps. Results are presented in \cref{fig:segmentation_sd}. From \cref{fig:segmentation_sd}, we see that the generated images show a clear separation between object and background that matches the given segmentation map nearly perfectly. The generated object and background also each match their descriptive text (i.e. dog breed and environment description). Furthermore, the generated images are overall highly realistic.

\paragraph{Face Recognition Guidance.}
To guide image generation to resemble the face of a given person, we compose a guidance function that combines a face detection module and a face recognition module. This setup produces a facial attribute embedding from an input face image. We use multi-task cascaded convolutional networks (MTCNN)~\cite{zhang2016mtcnn} as the face detection module, and use facenet~\cite{schroff2015facenet} as the face recognition module. The guidance function $f$ hence crops out the detected face and outputs a facial attribute embedding as prompt, while we use $l_1$-loss between embedding as the loss function $\ell$. We note that to compute the guidance direction in our algorithm, we only backpropagate through the facenet and treat the face cropping mask produced by MTCNN as an oracle input, as MTCNN utilizes non-maximum suppression~\cite{neubeck2006nms} which is non-differentiable. Here we set $s(t) = 20000 \cdot \sqrt{1 - \alpha_t}$ and and $k=2$.

\begin{table}[t]
\vspace{18pt}
\centering
\resizebox{\columnwidth}{!}{%
\begin{tabular}{cccc}
    {\fontsize{33}{40}\selectfont \backslashbox{Guide}{Prompt}} &
    
    \vtop{\vspace{-50pt}\hsize=\columnwidth \hangindent=0em \nextline{Headshot of a} \nextline{woman with }\nextline{a dog.}}&
    
    \vtop{\vspace{-50pt}\nextline{Headshot of a}\nextline{woman with a }\nextline{dog on beach.}}&
    
    \vtop{\vspace{-50pt}\nextline{An oil painting of}\nextline{a headshot of a}\nextline{women with a dog.}}\\
    \\
    \vtop{\nextline{(N/A)}} &
    \includegraphics[width=\gw,valign=m]{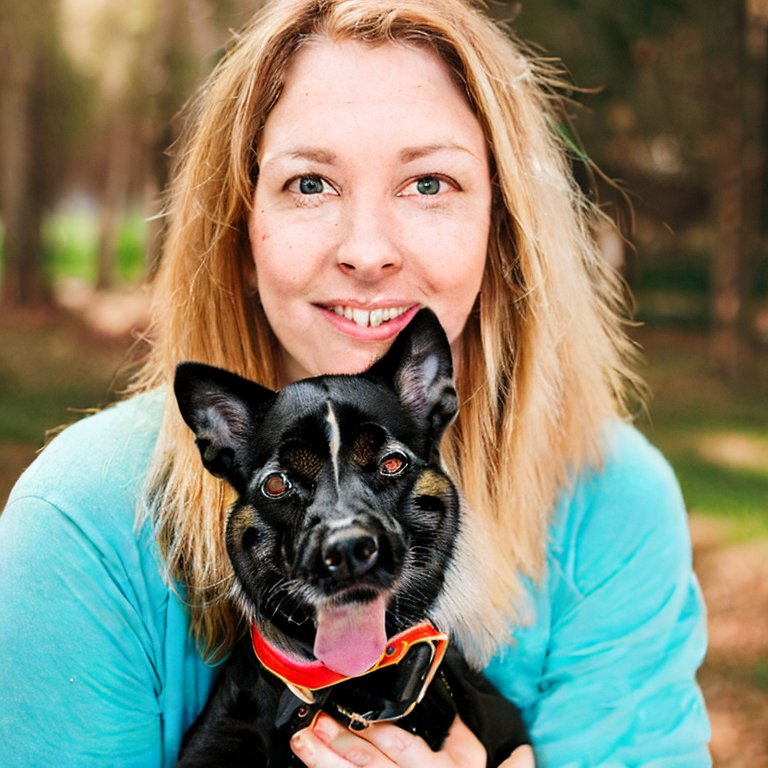} &
    \includegraphics[width=\gw,valign=m]{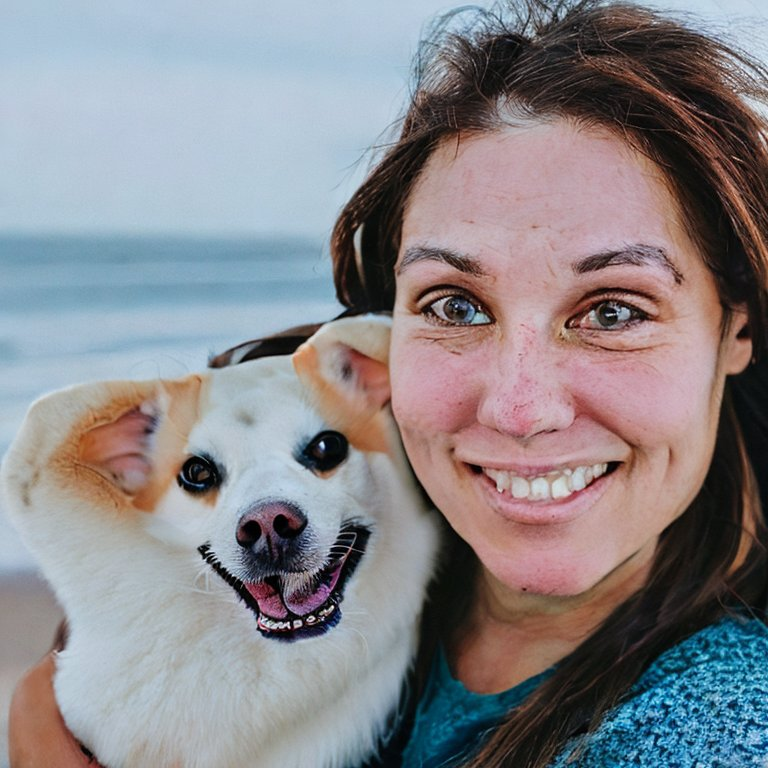} &
    \includegraphics[width=\gw,valign=m]{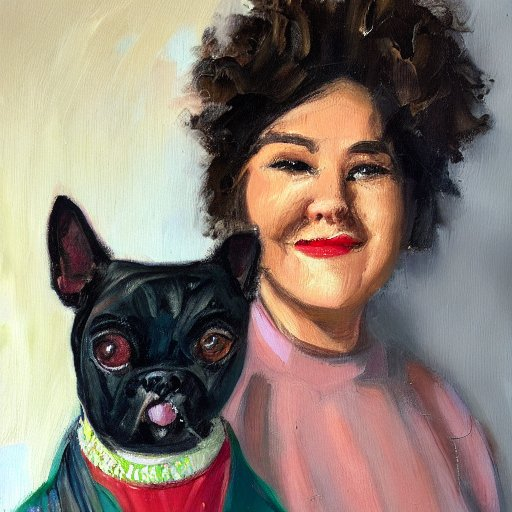}  \\

    \includegraphics[width=\gw,valign=m]{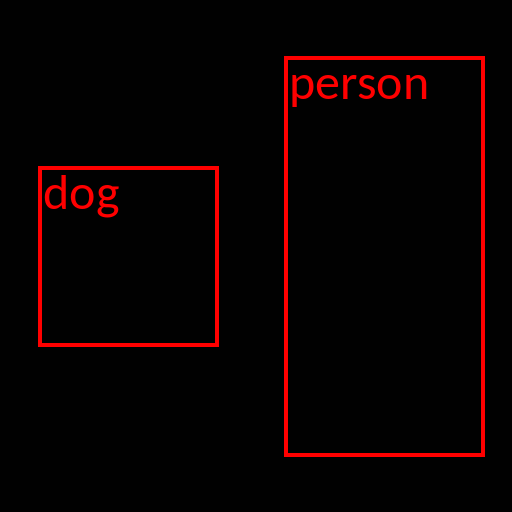} &
    \includegraphics[width=\gw,valign=m]{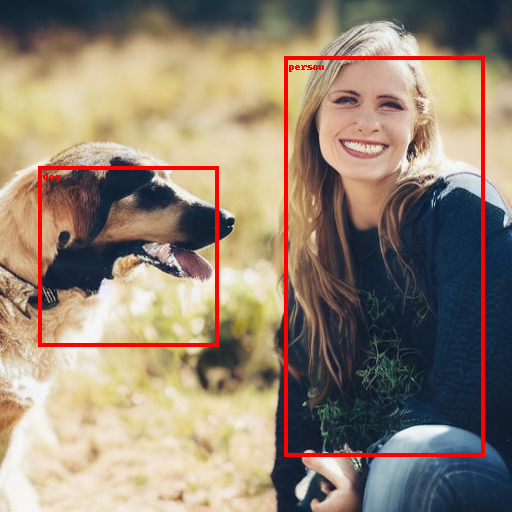} &
    \includegraphics[width=\gw,valign=m]{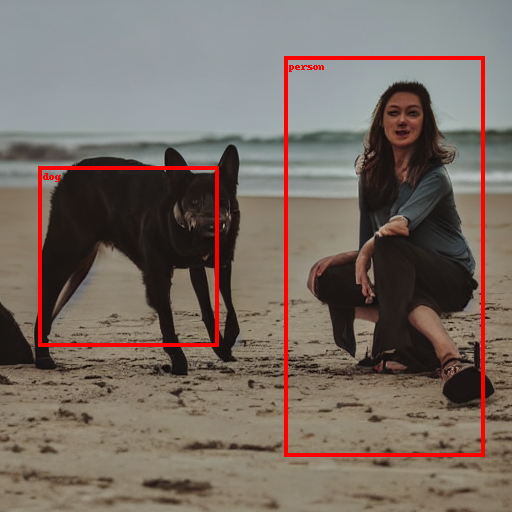} &
    \includegraphics[width=\gw,valign=m]{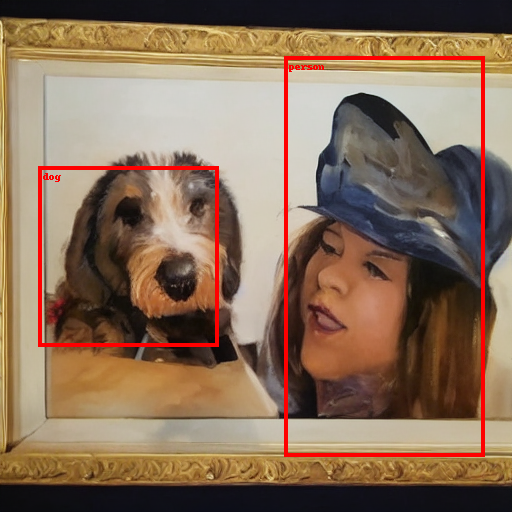} \\
    
    \includegraphics[width=\gw,valign=m]{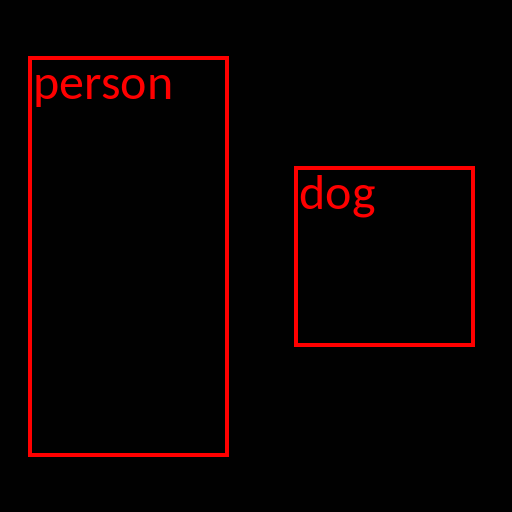} &
    \includegraphics[width=\gw,valign=m]{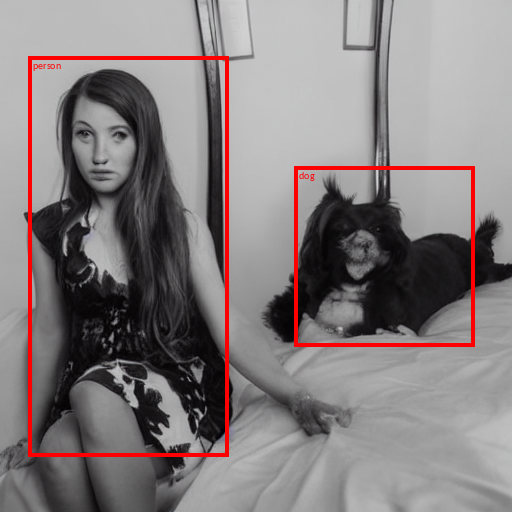} &
    \includegraphics[width=\gw,valign=m]{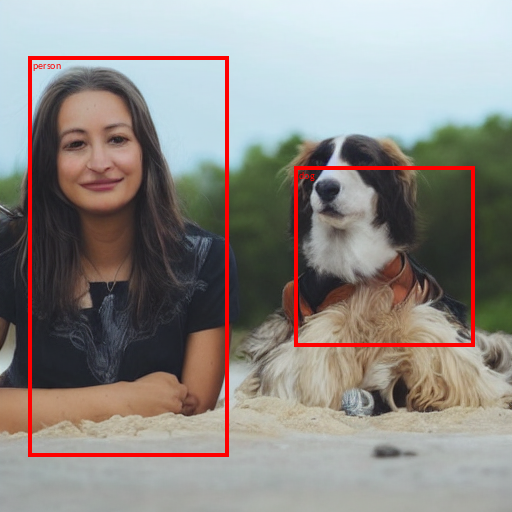} &
    \includegraphics[width=\gw,valign=m]{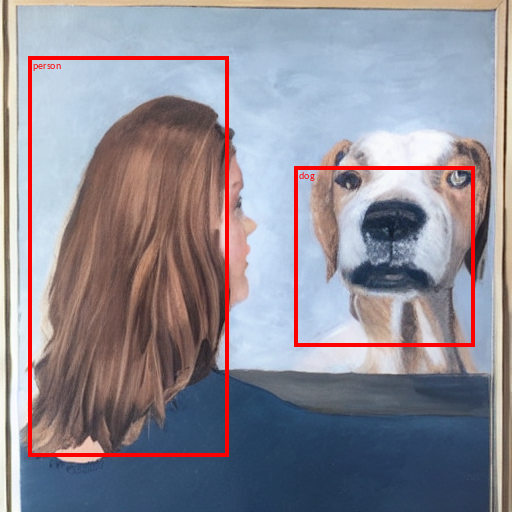} \\

    \includegraphics[width=\gw,valign=m]{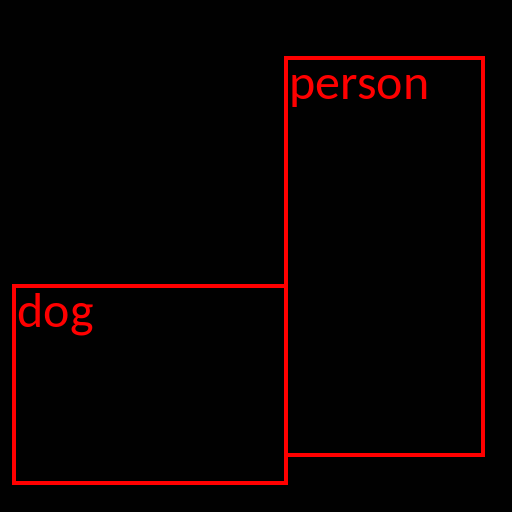} &
    \includegraphics[width=\gw,valign=m]{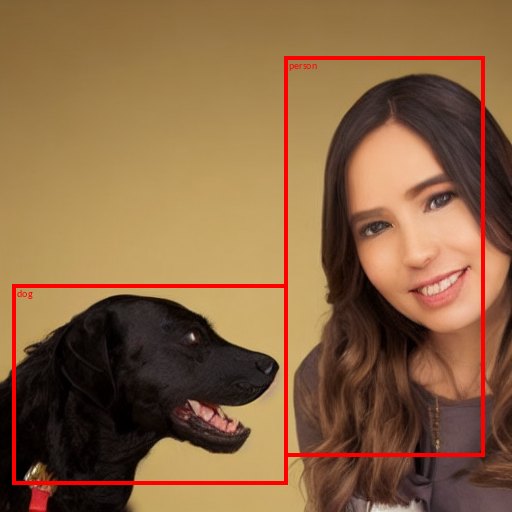} &
    \includegraphics[width=\gw,valign=m]{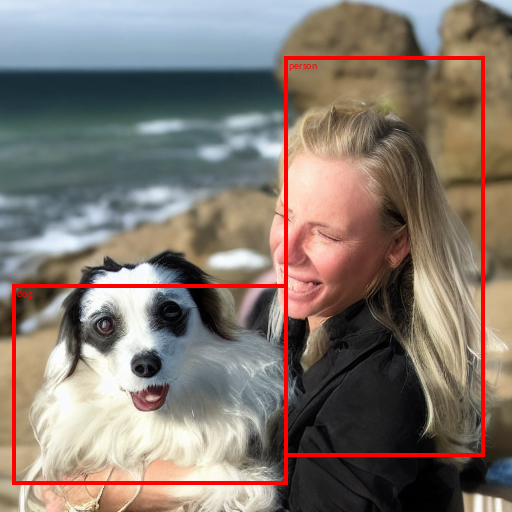} &
    \includegraphics[width=\gw,valign=m]{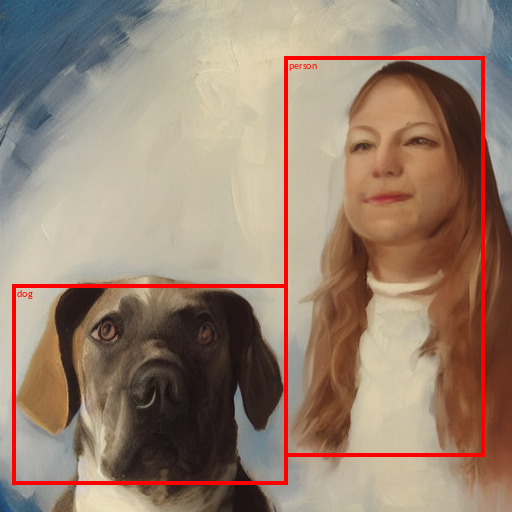} \\
    
\end{tabular}}
\captionof{figure}[]{In addition to matching the text prompts (above each column), these images are guided by an object detector. Each column contains examples of images generated to match the prompt and the bounding boxes used for guidance. The top row contains examples generated without guidance.}
\label{fig:od_sd}
\vspace{-.2cm}
\end{table}

\begin{table}[ht]
\vspace{18pt}
\centering
\resizebox{\columnwidth}{!}{%
\begin{tabular}{cccc}
    {\fontsize{33}{40}\selectfont \backslashbox{Style}{Prompt}} &
    
    \vtop{\vspace{-50pt}\hsize=\columnwidth \hangindent=0em \nextline{A colorful } \nextline{photo of an }\nextline{Eiffel Tower}}&
    
    \vtop{\vspace{-50pt}\nextline{A fantasy photo}\nextline{of volcanoes}}&
    
    \vtop{\vspace{-50pt}\nextline{A portrait of}\nextline{a woman}}\\
    \\
    \vtop{\nextline{(N/A)}} &
    \includegraphics[width=\gw,valign=m]{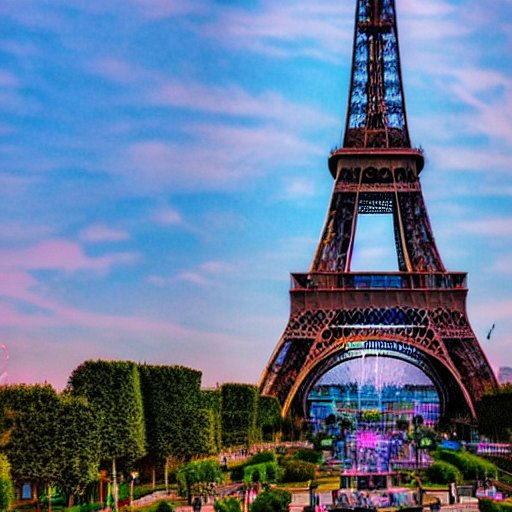} &
    \includegraphics[width=\gw,valign=m]{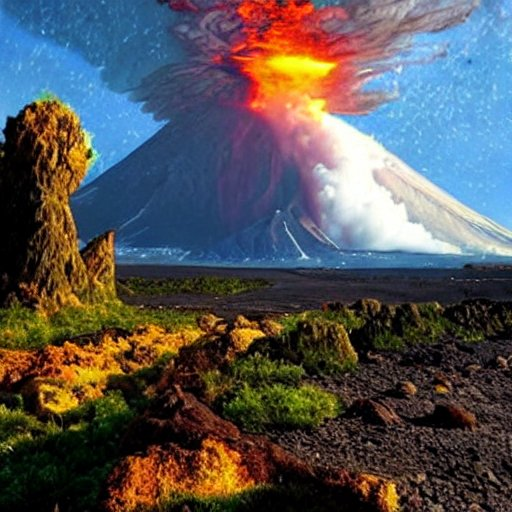} &
    \includegraphics[width=\gw,valign=m]{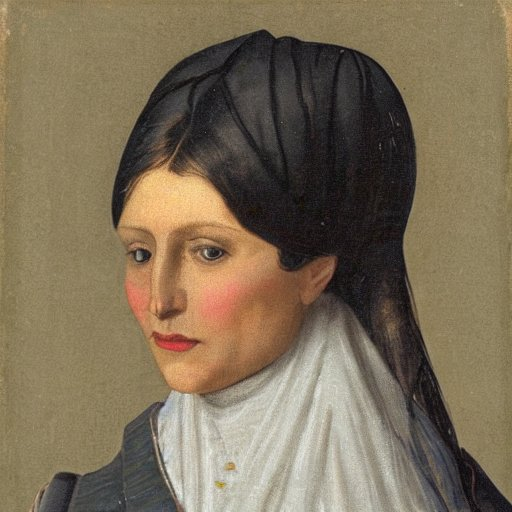}  \\

    \includegraphics[width=\gw,valign=m]{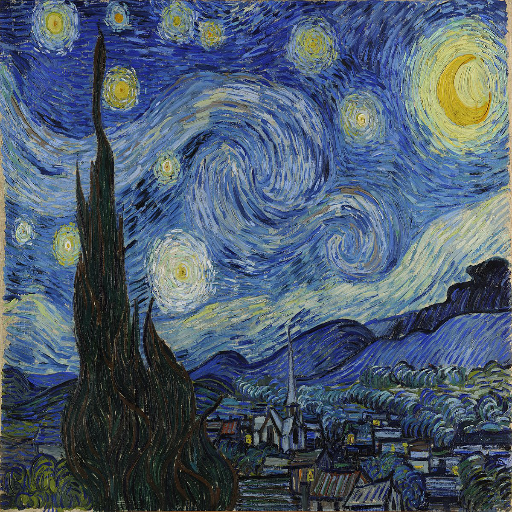} &
    \includegraphics[width=\gw,valign=m]{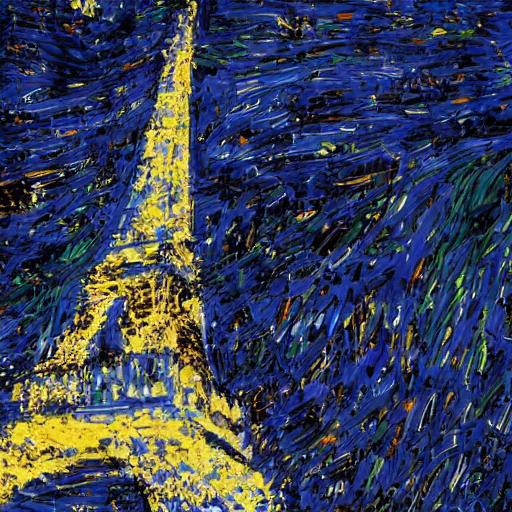} &
    \includegraphics[width=\gw,valign=m]{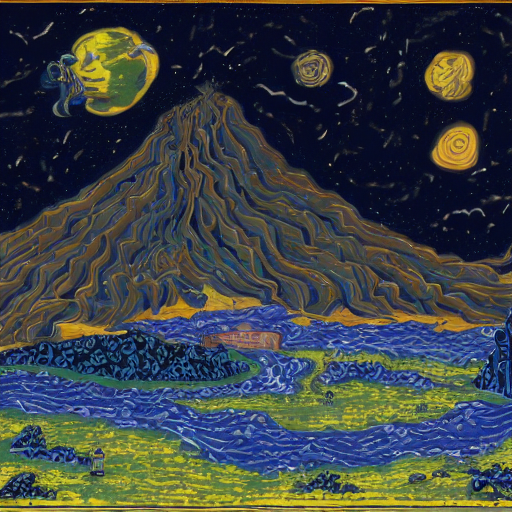} &
    \includegraphics[width=\gw,valign=m]{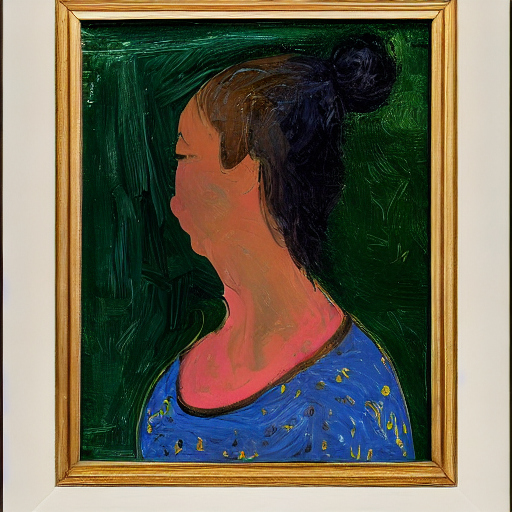}  \\

    \includegraphics[width=\gw,valign=m]{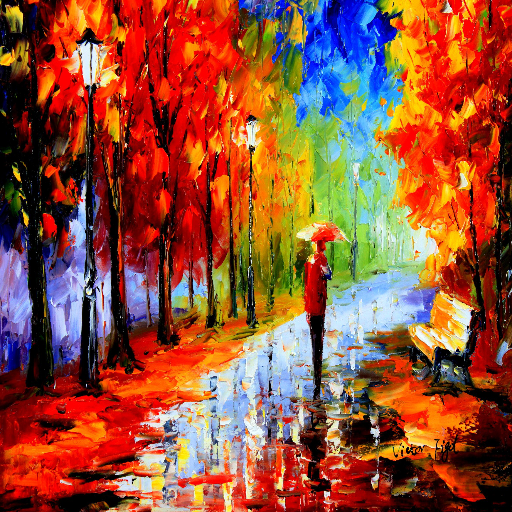} &
    \includegraphics[width=\gw,valign=m]{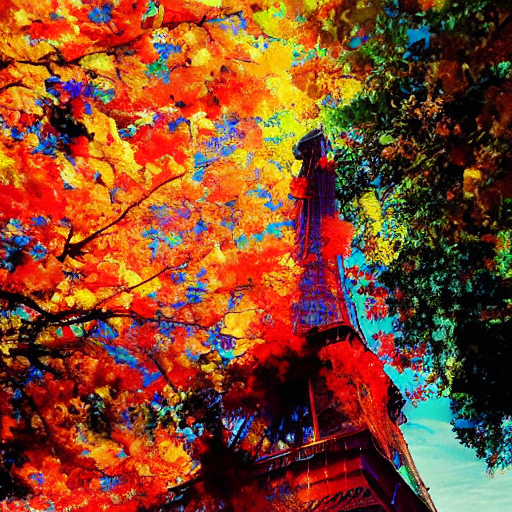} &
    \includegraphics[width=\gw,valign=m]{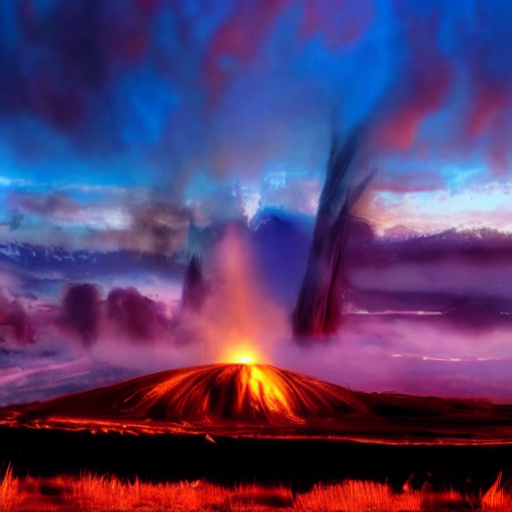} &
    \includegraphics[width=\gw,valign=m]{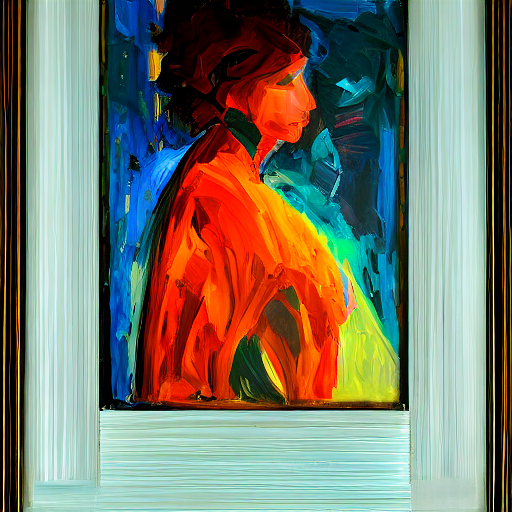}  \\

    \includegraphics[width=\gw,valign=m]{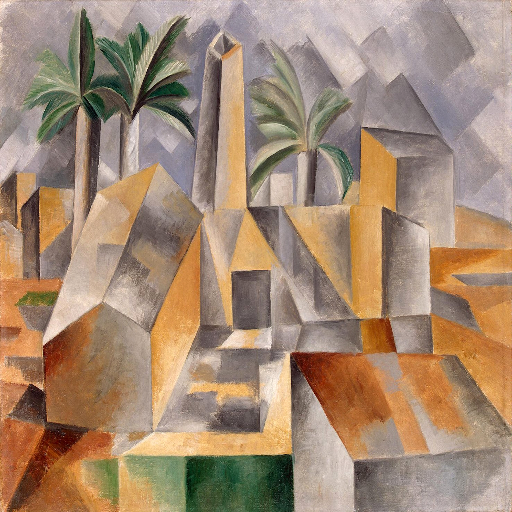} &
    \includegraphics[width=\gw,valign=m]{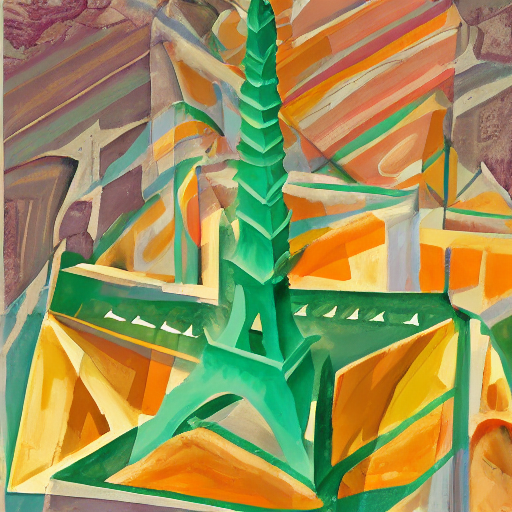} &
    \includegraphics[width=\gw,valign=m]{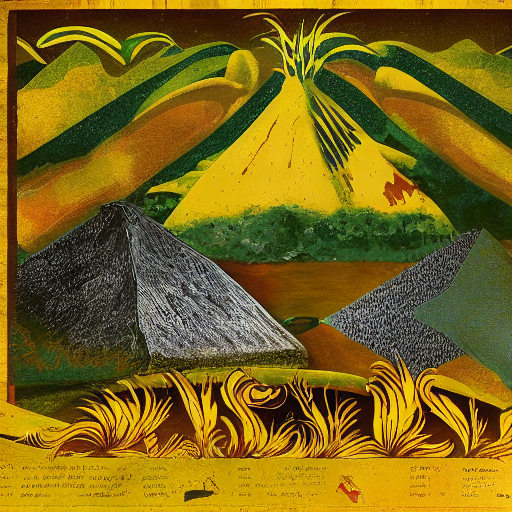} &
    \includegraphics[width=\gw,valign=m]{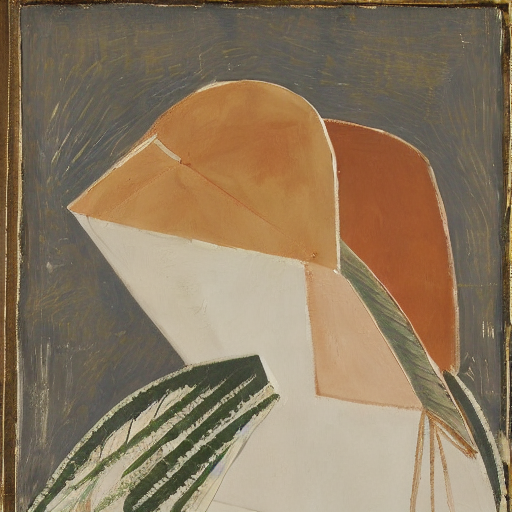}  \\
    
\end{tabular}}
\captionof{figure}[]{In addition to matching the text prompts (above each column), these images are guided by a style image. Each column contains examples of images generated to match the text prompt and the style image used for guidance. The top-most row contains examples generated without style guidance.}
\label{fig:st_sd}
\vspace{-.2cm}
\end{table}

We explore different combinations of face guidance and text prompts. Similarly to the segmentation case, we use the text prompt as a fixed additional conditioning to Stable Diffusion and guide this text-conditional trajectory with our algorithm so that the face in the generated image looks similar to the face prompt. In \cref{fig:face_sd}, we clearly see that the facial characteristics of a given face prompt are reproduced almost perfectly on the generated images. The descriptive text of either background, material, or style is also realized correctly and blends nicely with the generated faces. 

\begin{figure}[t]
    \centering
    \includegraphics[width=\columnwidth]{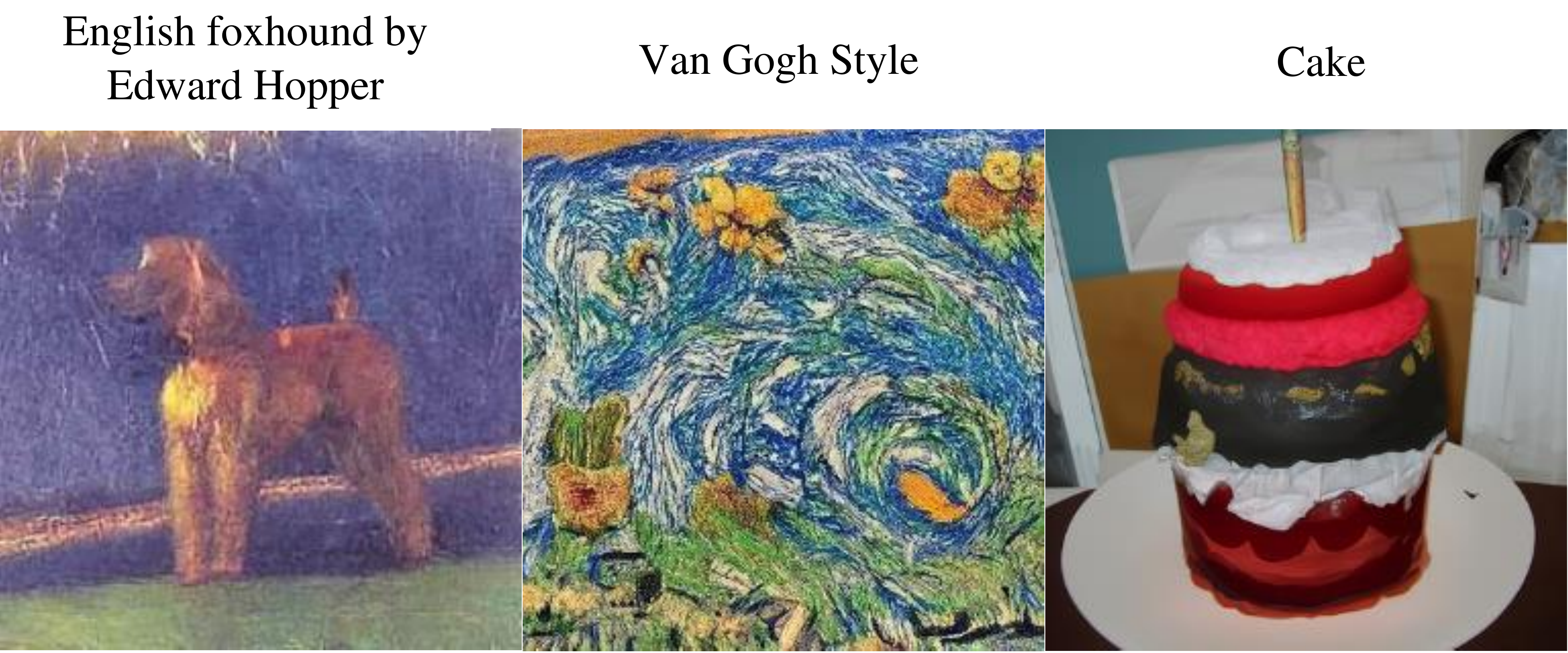}
    \caption{We show that unconditional diffusion models trained on ImageNet can be guided with CLIP to generate high-quality images that match the text prompts, even if these generated images should be \textit{out of distribution}. 
    }
    \label{fig:ood_imagenet}
    \vspace{-.2cm}
\end{figure}

\vspace{-.1cm}
\paragraph{Object Location Guidance}
For Stable Diffusion, we also present the results  guiding image generation with an object detection network. For this experiment, we use Faster-RCNN~\cite{ren2015faster_rcnn} with Resnet-50-FPN backbone~\cite{li2021resent50fpn}, a \href{https://pytorch.org/vision/stable/models/generated/torchvision.models.detection.fasterrcnn_resnet50_fpn_v2.html#torchvision.models.detection.fasterrcnn_resnet50_fpn_v2}{publicly available} pre-trained model in Pytorch, as our object detector. We use bounding boxes with class labels as our object location prompt. We construct a loss function $\ell$ by the sum of three individual losses, namely (1) anchor classification loss, (2) bounding box regression loss and (3) region label classification loss, where (1) and (2) are computed on the region proposal head while (3) is computed on the region classification head. We note that, compared to standard R-CNN training, we drop the additional bounding box alignment loss on region classification head. We found that our loss construction helps to produce objects of correct categories for each location prompt. We set $s(t) = 100 \cdot \sqrt{1 - \alpha_t}$ and $k=3$.

We again experiment with different combinations of text prompt and object location prompt, and similarly use the text prompt as a fixed  conditioning to Stable Diffusion. Using our proposed guidance algorithm, we perform guided image generation that generates and matches the objects presented in the text prompt to the given object locations. The results are presented in \cref{fig:od_sd}. We observe from \cref{fig:od_sd} that objects in the descriptive text all appear in the designated location with the appropriate size indicated by the given bounding boxes. Each location is filled with appropriate, high-quality generations that align with varied image content prompts, ranging from ``beach" to ``oil painting". 


\paragraph{Style Guidance}
Finally, we conclude our experiments on Stable Diffusion by guiding the image generation based on a reference style given by a style image. To achieve so, we capture the reference style from the style image by the image feature extractor from CLIP, and use the resulting image embedding as prompts. 
The loss function calculates the negative cosine similarity between the embedding of generated images and the embedding of the style image.
Similar to previous experiments, we control the content using text input as additional conditioning to the Stable Diffusion model. 

We experiment with combinations of different style images and different text prompts, and present the results in \cref{fig:st_sd}. From \cref{fig:st_sd}, we can see that the generated images contain contents that match the given text prompts, while exhibiting style that matches the given style images.
In this experiment we set $s(t) = 6 \cdot \sqrt{1 - \alpha_t}$ and $k=6$. Furthermore, in order to control the amount of content we set the scale $\gamma$, a parameter of Stable Diffusion that balances the text-conditional generation and unconditional generation, as 3.0, 3.0, and 4.0 respectively for each column. 


\subsection{Results for ImageNet Diffusion}
\label{sec:exp:imagenet}
In this section, we present results for guided image generation using an unconditional diffusion model trained on ImageNet. We experiment with CLIP guidance, object location guidance and a hybrid guided image generation task which we term segmentation-guided inpainting. We will discuss results and implementations of each guidance in its corresponding subsection. 
\paragraph{CLIP Guidance.}
We use the same construction of $f$ and $\ell$ for Stable Diffusion to perform CLIP-guided generation. We use only forward guidance for this experiment.
To assess the limit of our universal guidance algorithm, we hand-crafted text prompts such that the matching images are \textit{expected to be out of distribution}. In particular, our text prompts either designate art styles that are far from realistic or designate objects that do not belong to any possible class label of ImageNet. We present the results in \cref{fig:ood_imagenet}, and from the results, we clearly see that our algorithm still successfully guides the generation to produce quality images that also match the text prompts. For all three images, we have $s(t) = w \cdot \sqrt{1-\alpha_t}$, where $w$ is 2, 5 and 2 respectively and $k$ is 10, 5 and 10 respectively.

\begin{table}[t]
\centering
\resizebox{\columnwidth}{!}{%
\begin{tabular}{ccc}
    \fontsize{22}{22}\selectfont Object Location & \fontsize{22}{22}\selectfont Forward Only &  \fontsize{22}{22}\selectfont Forward + Backward \\ 

    \includegraphics[width=\gw,valign=m]{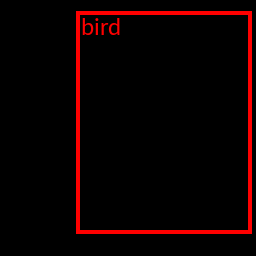} &
    \includegraphics[width=\gw,valign=m]{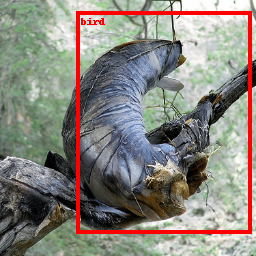} & \includegraphics[width=\gw,valign=m]{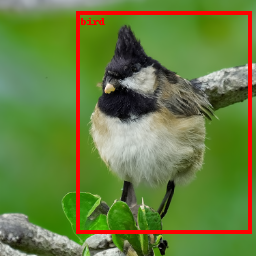}\\
    \\
    
    \includegraphics[width=\gw,valign=m]{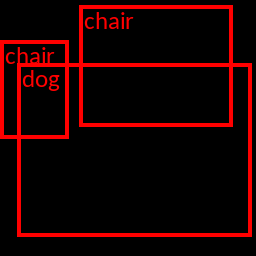} &
    \includegraphics[width=\gw,valign=m]{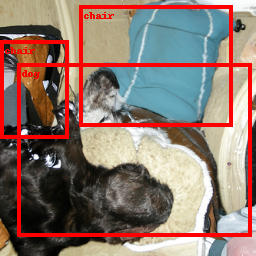} & \includegraphics[width=\gw,valign=m]{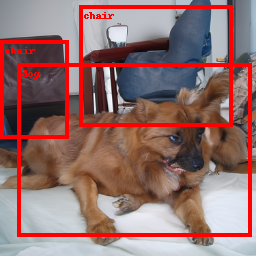} \\
     \\
\end{tabular}}
\captionof{figure}[]{Generation guided by object detection with the unconditional ImageNet model. Images generated with both forward and backward guidance are realistic and have the desired objects in the designated locations. In contrast, images generated using only forward guidance exhibit objects of the incorrect category or with inaccurate position/size.
}
\label{fig:od_imagenet}
\vspace{-.4cm}
\end{table}

\paragraph{Object Location Guidance.}
Similar to object location guidance for Stable Diffusion, we also use the same network architecture and the same pre-trained model as our object detection network, and construct an identical loss function $\ell$ for our guidance algorithm. However, unlike Stable Diffusion, object locations are the only prompts available for guided image generation. For this experiment, we use $s(t) = 100 \sqrt{1-\alpha_t}$ and $k = 3$.

We again experiment with different object location prompts using two configurations of our algorithm, namely (1) using only forward universal guidance and (2) using both forward and backward universal guidance. 
We observe from \cref{fig:ood_imagenet} that applying both forward and backward guidance generates images that are realistic and the objects matches the prompt nicely. On the other hand, while images generated using only forward guidance remain realistic, they feature objects with mismatching categories and locations. 
The results demonstrate the effectiveness of our universal guidance algorithm, and also validate the necessity of our backward guidance.

\begin{table}[t]
\centering
\resizebox{\columnwidth}{!}{%
\begin{tabular}{ccc}
    \fontsize{24}{22}\selectfont Masked Image & \fontsize{24}{22}\selectfont Clf. Guided & \fontsize{24}{22}\selectfont Clf. + Seg. Guided \\

    \includegraphics[width=\gw,valign=m]{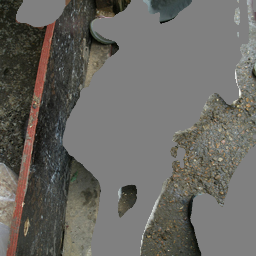} &
    \includegraphics[width=\gw,valign=m]{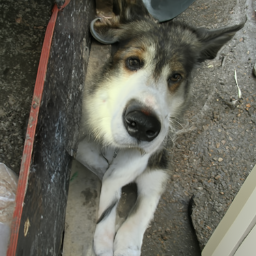} & \includegraphics[width=\gw,valign=m]{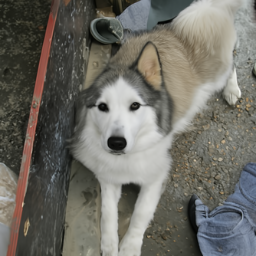}\\

    \includegraphics[width=\gw,valign=m]{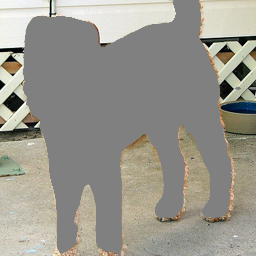} &
    \includegraphics[width=\gw,valign=m]{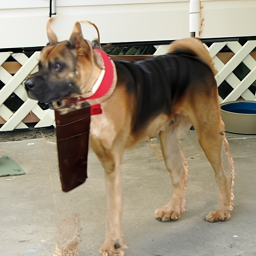} & \includegraphics[width=\gw,valign=m]{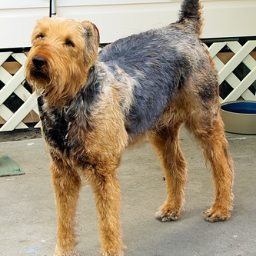}\\
    
     \\
\end{tabular}}
\captionof{figure}[]{Our guidance algorithm can incorporate feedback from multiple guidance functions. The first column shows the prompt for inpainting. The second column shows classifier-guided inpainting, where dog images with close matches to inpainting prompt are generated. The third column shows images generated with both classifier and segmentation guidance, where realistic dogs are generated exactly on the masked regions. The results show that our algorithm handles multiple guidance functions effectively.}
\label{fig:segmentation_guided_inpainting}
\vspace{-0.3cm}
\end{table}

\paragraph{Segmentation-Guided Inpainting.}
In this experiment, we aim to explore the ability of our algorithm to handle multiple guidance functions. We perform guided image generation with combined guidance from an inpainting mask, a classifier and a segmentation network.
We first generate images with masked regions as the prompt for inpainting.
We then pick an object class $c$ as the prompt for classification and generate a segmentation mask where the masked regions are considered foreground objects of the same class $c$.
We use $\ell_2$ loss on the non-masked region as the loss function for inpainting, and set the corresponding $s(t) = 0$, or equivalently only use backward guidance for inpainting.
We use the same segmentation network as described in \cref{sec:exp:stable_diff} with $s(t) = 200\sqrt{1-\alpha_t}.$
For classification guidance, we use the classifier that accepts noisy input~\cite{dhariwal21diffusion_beats_gan}, and perform the original classifier guidance \cref{eq:classifier_guidance_original} instead of our forward guidance.
The results summarized in \cref{fig:segmentation_guided_inpainting} show that when using both inpainting and classifier as guidance, our algorithm generates realistic images that both match the inpainting prompt and can be classified correctly to the given object class. 
Adding in segmentation guidance, our algorithm further improves the generated images with a near-perfect match to both the segmentation map and inpainting prompt while maintaining realism. 
This demonstrates that our algorithm can effectively combine the feedback from individual guidance functions.


\section{Limitations}
\label{sec:limitation}
  Generation using universal guidance is typically slower than standard conditional generation for several reasons.  Empirically, multiple iterations of denoising are required at every noise level $t$ to generate high-quality images with complex guidance functions. However, the time complexity of our algorithm scales linearly with the number of recurrence steps $k$, which slows down image generation when $k$ is large.  Also, as demonstrated in the main paper, backward guidance is required in certain scenarios to help generate images that match the given constraint. Computing backward guidance requires performing minimization with a multi-step gradient descent inner loop. While proper choices of gradient-based optimization algorithms and learning rate schedules significantly speed up the convergence of minimization, the time it takes to compute backward guidance inevitably becomes longer when the guidance function is itself a very-large neural network.  Finally, we note that, to get optimal results, sampling hyper-parameters must be chosen individually for each guidance network.

\section{Conclusion}
In this paper, we propose a universal guidance algorithm that is able to perform guided image generation with any off-the-shelf guidance function based on a fixed foundation diffusion model. Our algorithm only requires guidance and loss functions to be differentiable, and avoids any retraining to adapt either the guidance function or the foundation model to a specific type of prompt. We demonstrate promising results with our algorithm on complex guidance including segmentation, face recognition and object detection systems. Even multiple guidance functions can be combined and used in conjunction. 

\section{Acknowledgements}
This work was made possible by the National Science Foundation (IIS-2212182), the AFOSR MURI Program, the Office of Naval Research (N000142112557), the ONR MURI program, IARPA WRIVA, and Capital One Bank.

\bibliography{main}
\bibliographystyle{icml2023}

\newpage
\appendix
\onecolumn

\section{More results}
\label{appendix:more_results}

\begin{figure}[htp]
    \centering
    \begin{subfigure}[b]{\textwidth}
        \includegraphics[width=\textwidth]{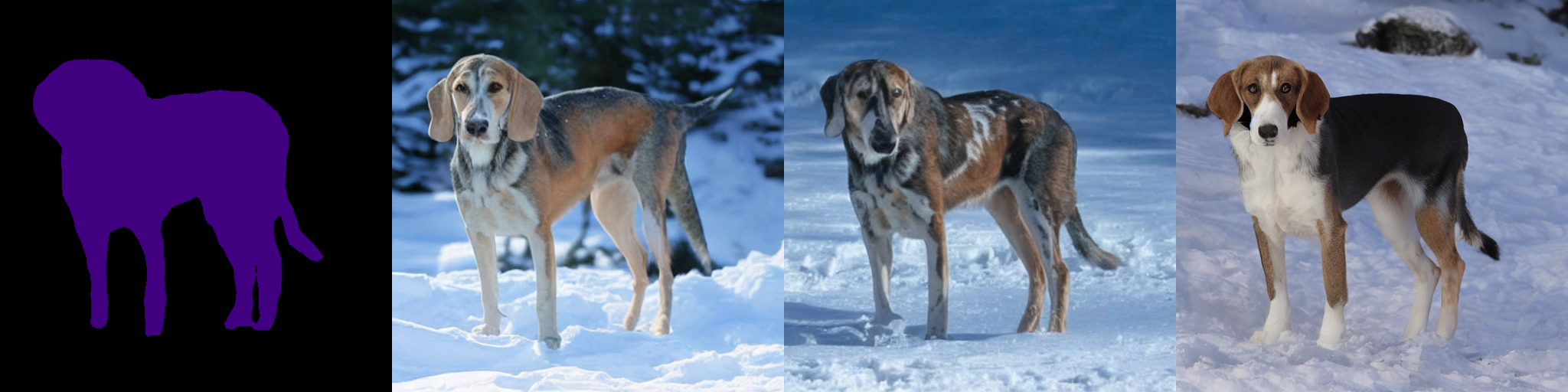}
        \caption{Walker hound, Walker foxhound on snow.}
    \end{subfigure}
    \begin{subfigure}[b]{\textwidth}
        \includegraphics[width=\textwidth]{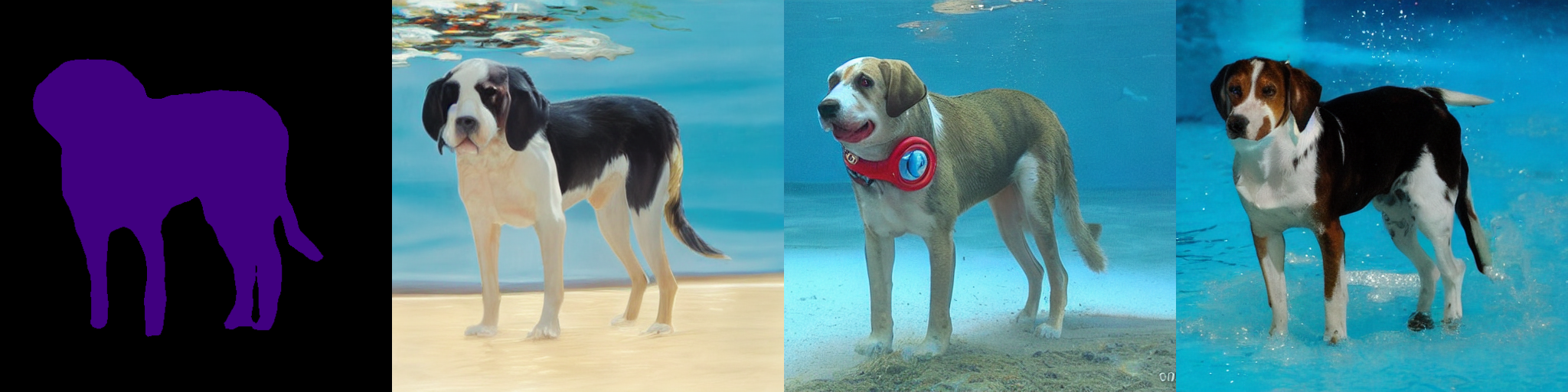}
        \caption{Walker hound, Walker foxhound under water.}
    \end{subfigure}
    \begin{subfigure}[b]{\textwidth}
        \includegraphics[width=\textwidth]{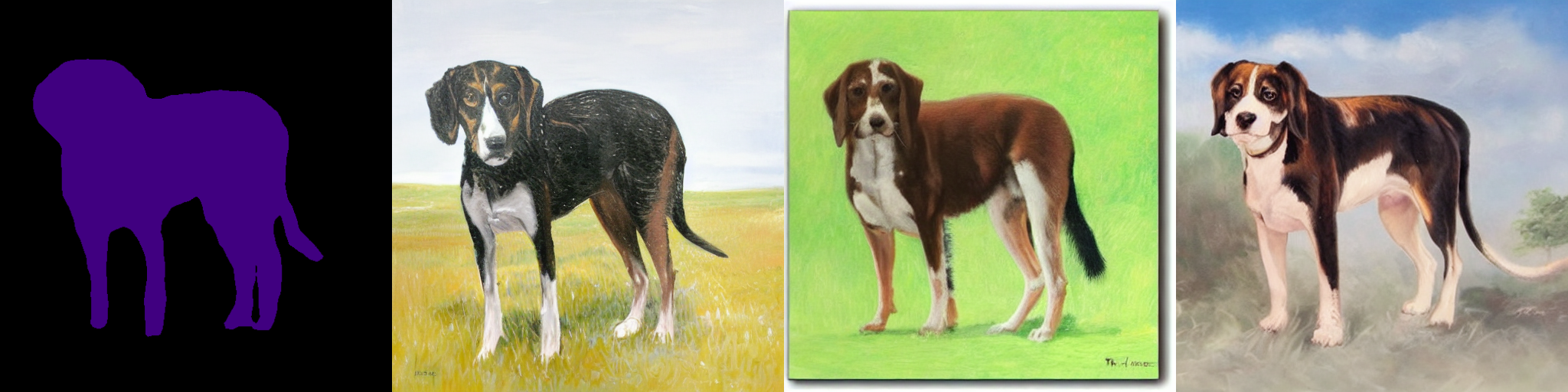}
        \caption{Walker hound, Walker foxhound as an oil painting.}
    \end{subfigure}
    \caption{More images to show Segmentation guidance. In each subfigure, the first image is the segmentation map used to guide the image generation with its caption as its text prompt.}
    \label{fig:app_seg}
\end{figure}

\begin{figure}[htp]
    \centering
    \begin{subfigure}[b]{\textwidth}
        \includegraphics[width=\textwidth]{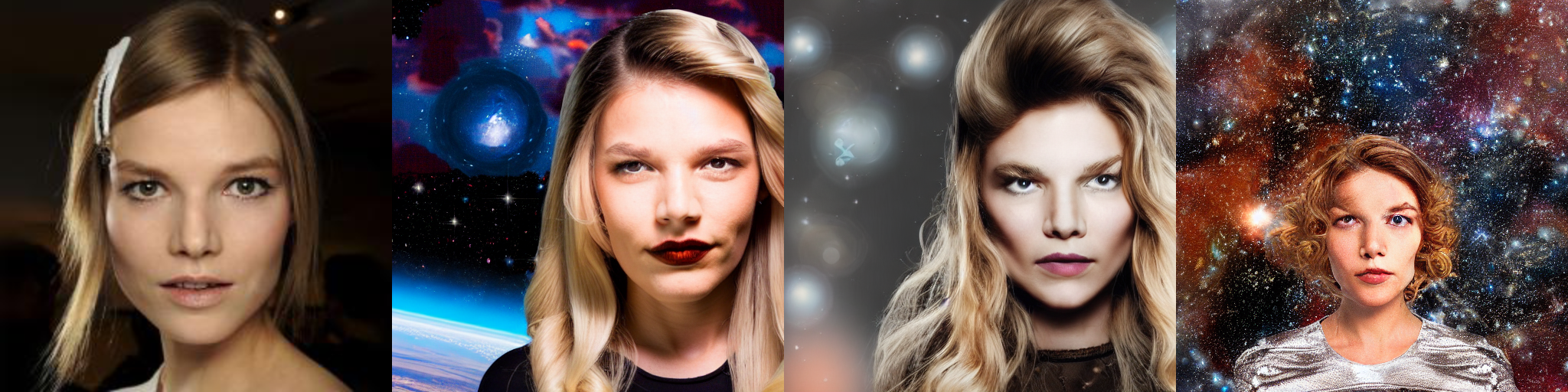}
        \caption{Headshot of a person with blonde hair with space background.}
    \end{subfigure}
    \begin{subfigure}[b]{\textwidth}
        \includegraphics[width=\textwidth]{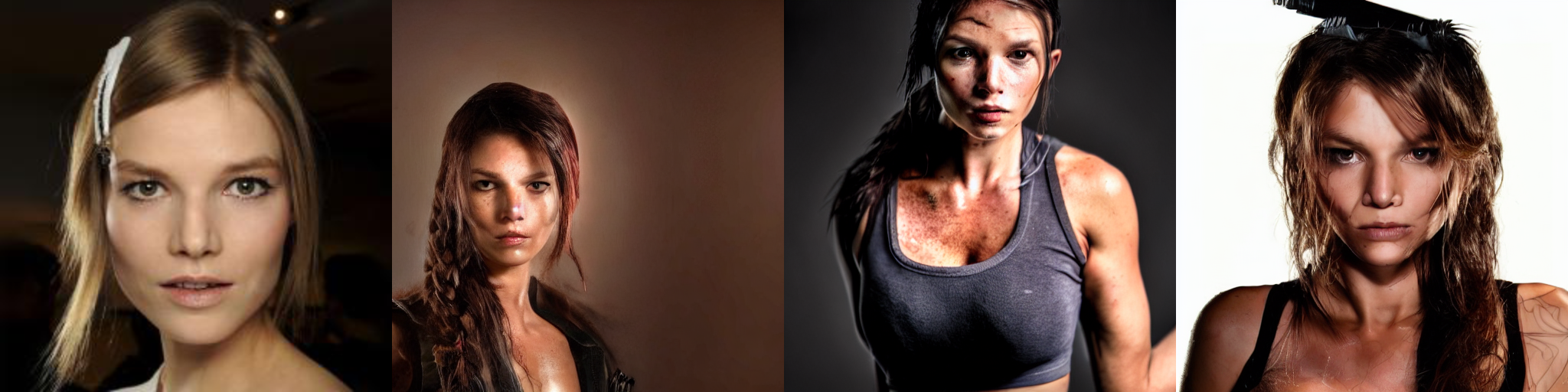}
        \caption{A headshot of a woman looking like a lara croft.}
    \end{subfigure}
    \begin{subfigure}[b]{\textwidth}
        \includegraphics[width=\textwidth]{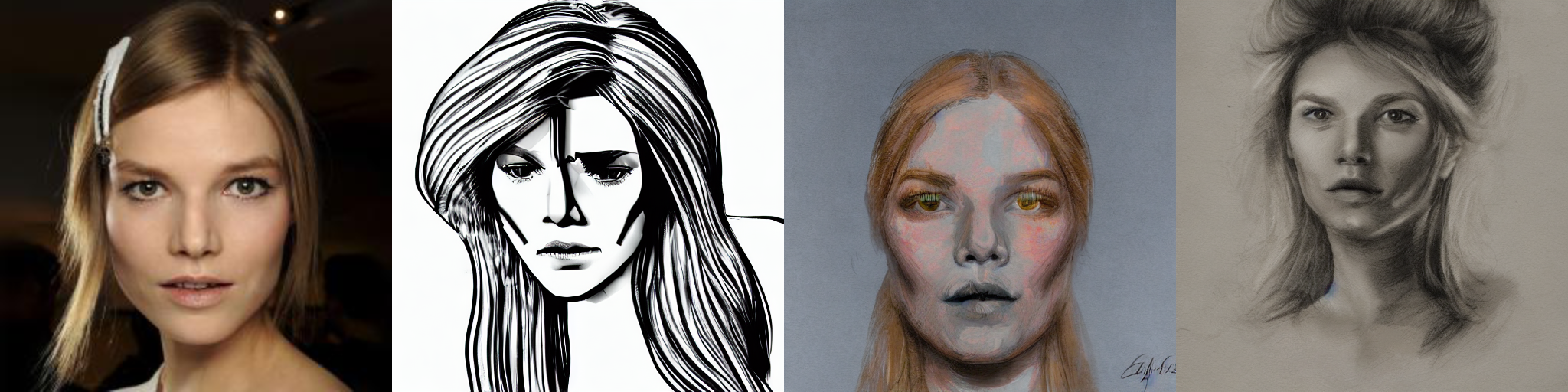}
        \caption{A headshot of a blonde woman as a sketch.}
    \end{subfigure}
    \caption{More images to show Face guidance. In each subfigure, the first image is the human identity used to guide the image generation with its caption as its text prompt.}
    \label{fig:app_face}
\end{figure}

\begin{figure}[htp]
    \centering
    \begin{subfigure}[b]{\textwidth}
        \includegraphics[width=\textwidth]{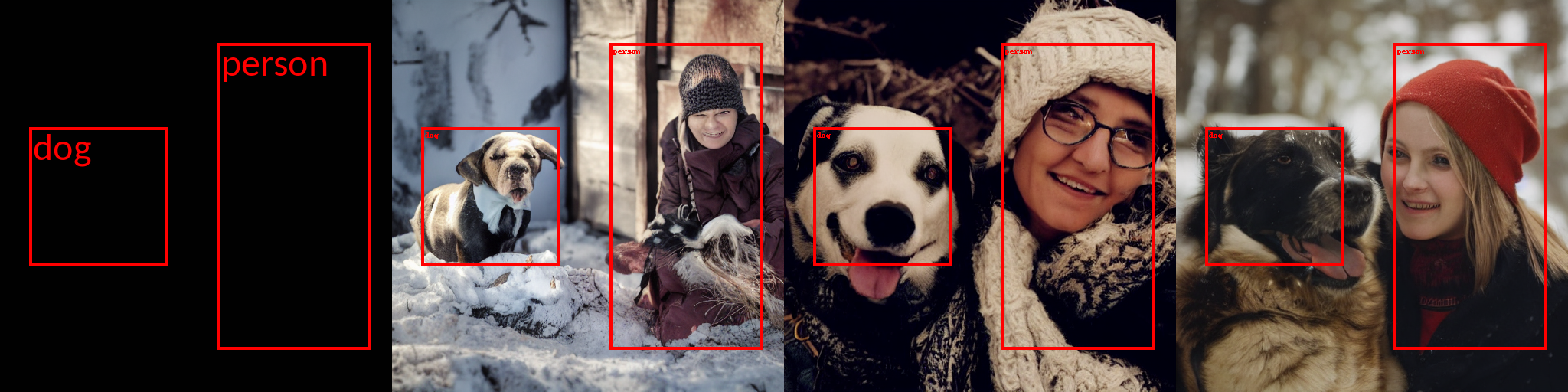}
        \caption{A headshot of a woman with a dog in winter.}
    \end{subfigure}
    \begin{subfigure}[b]{\textwidth}
        \includegraphics[width=\textwidth]{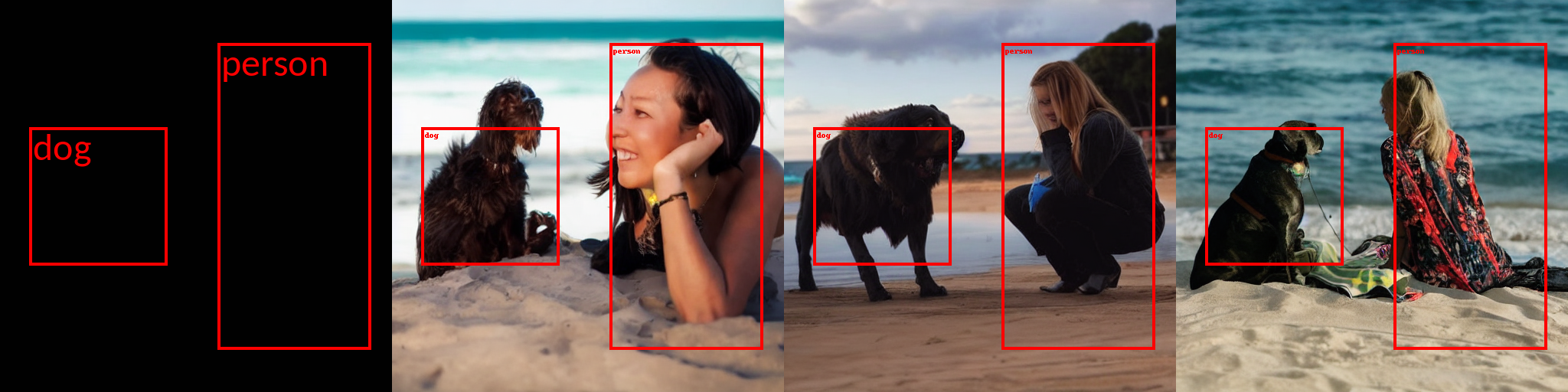}
        \caption{a headshot of a woman with a dog on beach.}
    \end{subfigure}
    \begin{subfigure}[b]{\textwidth}
        \includegraphics[width=\textwidth]{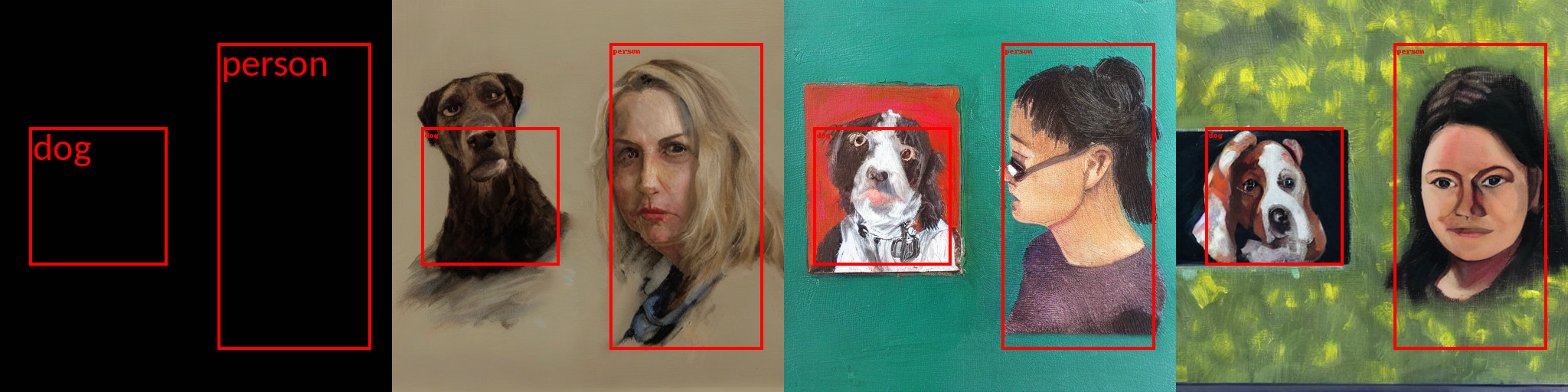}
        \caption{An oil painting of a headshot of a woman with a dog.}
    \end{subfigure}
    \caption{More images to show Object Location guidance. In each subfigure, the first image is the object location used to guide the image generation with its caption as its text prompt.}
    \label{fig:app_od}
\end{figure}

\begin{figure}[htp]
    \centering
    \begin{subfigure}[b]{\textwidth}
        \includegraphics[width=\textwidth]{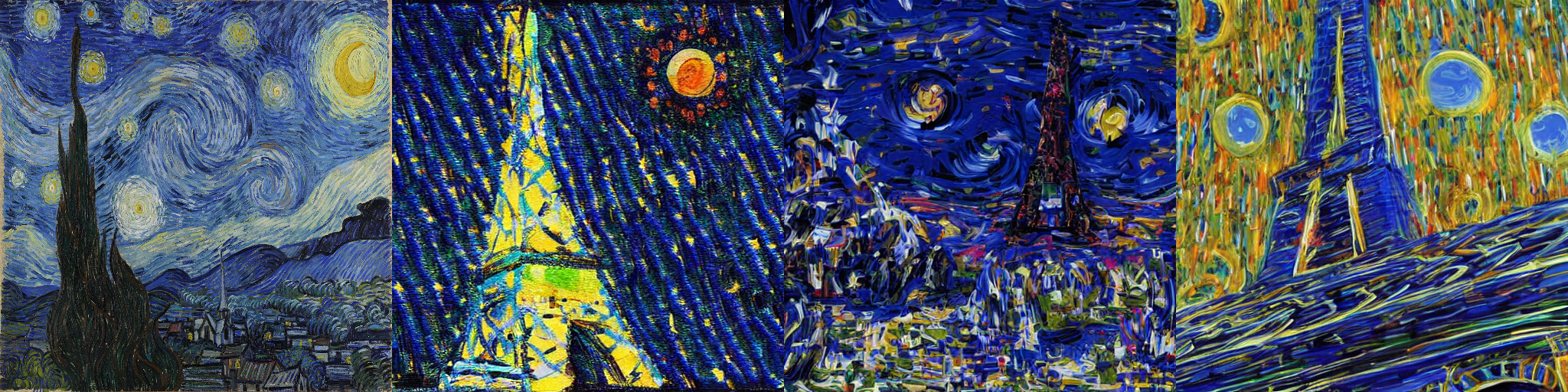}
        \caption{A colorful photo of an Eiffel Tower.}
    \end{subfigure}
    \begin{subfigure}[b]{\textwidth}
        \includegraphics[width=\textwidth]{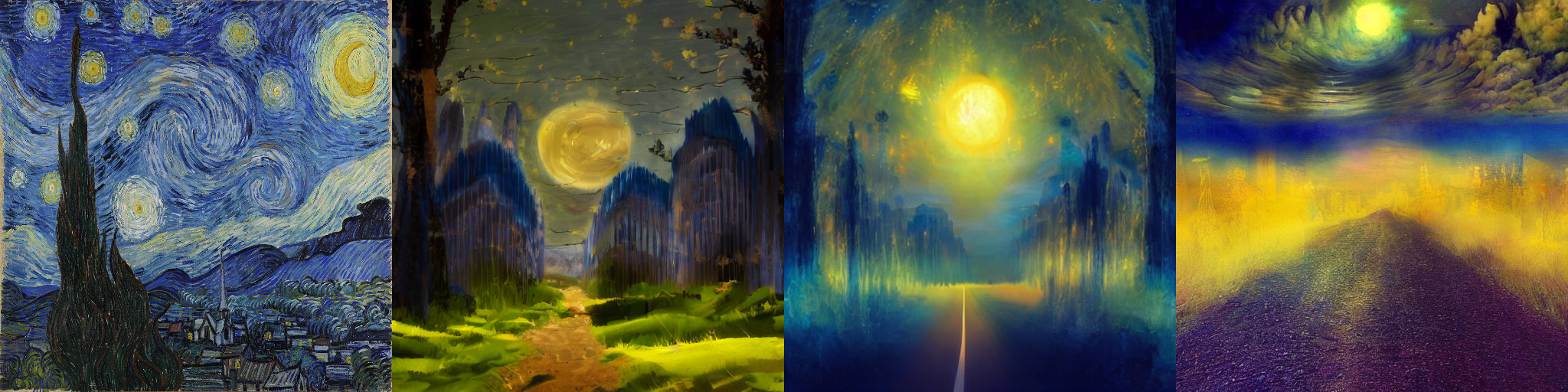}
        \caption{A fantasy photo of a lonely road.}
    \end{subfigure}
    \begin{subfigure}[b]{\textwidth}
        \includegraphics[width=\textwidth]{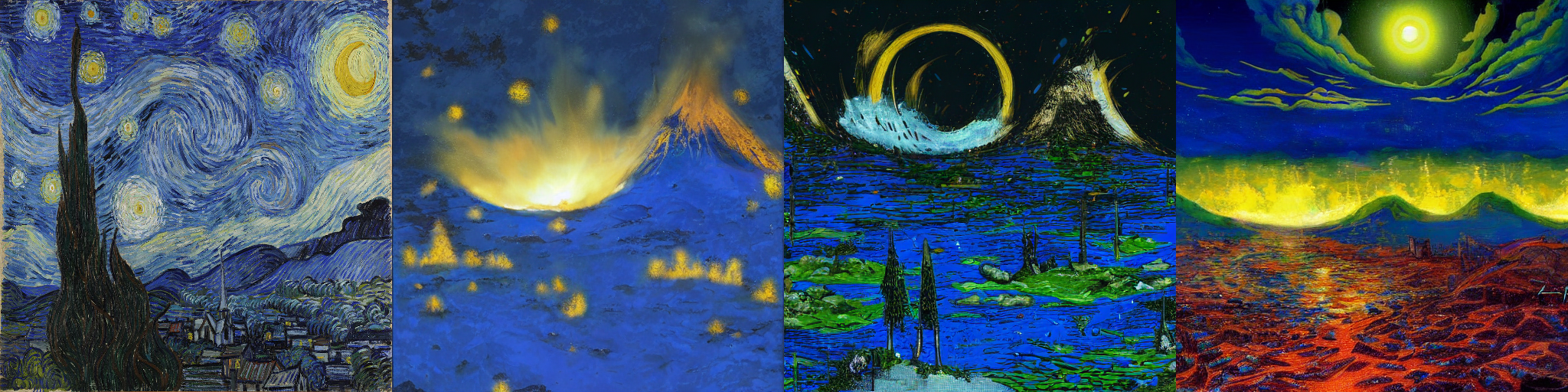}
        \caption{A fantasy photo of volcanoes.}
    \end{subfigure}
    \caption{More images to show Style Transfer. In each subfigure, the first image is the styling image used to guide the image generation with its caption as its text prompt.}
    \label{fig:app_st}
\end{figure}

\end{document}